\documentclass[runningheads]{llncs}

\usepackage{eccv}
\usepackage{eccvabbrv}

\usepackage{graphicx}
\usepackage{booktabs}

\usepackage[accsupp]{axessibility}  %
\usepackage{multirow} 
\usepackage{colortbl}
\usepackage{xcolor}
\usepackage{algorithm}
\usepackage{algpseudocode}

\usepackage{hyperref}

\usepackage{orcidlink}
\newcommand*{\ourmodel}{SLAck\@\xspace}

\begin{document}

\title{SLAck: Semantic, Location, and Appearance Aware Open-Vocabulary Tracking} 
\titlerunning{SLAck}

\author{
  Siyuan Li\inst{1} \and
  Lei Ke \inst{1} \and
  Yung-Hsu Yang\inst{1} \and
  Luigi Piccinelli \inst{1} \and
  Mattia Segù \inst{1} \and 
  Martin Danelljan \inst{1} \and
  Luc Van Gool \inst{1,2}
}
\authorrunning{S. Li et al.}
\institute{Computer Vision Lab, ETH Z{\"u}rich \and
INSAIT\\
}

\maketitle

\begin{abstract}
Open-vocabulary Multiple Object Tracking (MOT) aims to generalize trackers to novel categories not in the training set. Currently, the best-performing methods are mainly based on pure appearance matching. Due to the complexity of motion patterns in the large-vocabulary scenarios and unstable classification of the novel objects, the motion and semantics cues are either ignored or applied based on heuristics in the final matching steps by existing methods. In this paper, we present a unified framework SLAck that jointly considers semantics, location, and appearance priors in the early steps of association and learns how to integrate all valuable information through a lightweight spatial and temporal object graph. Our method eliminates complex post-processing heuristics for fusing different cues and boosts the association performance significantly for large-scale open-vocabulary tracking. Without bells and whistles, we outperform previous state-of-the-art methods for novel classes tracking on the open-vocabulary MOT and TAO TETA benchmarks.  Our code is available at \href{https://github.com/siyuanliii/SLAck}{github.com/siyuanliii/SLAck}.

\end{abstract}

\section{Introduction}

\begin{figure}[t]
  \centering
\includegraphics[width=1\linewidth]{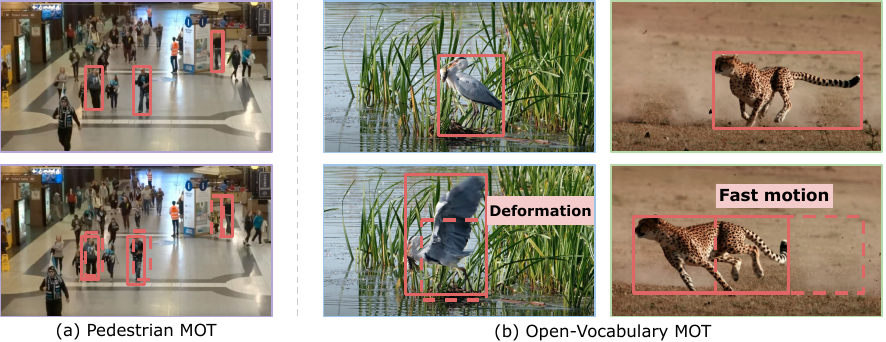}
  \caption{Unlike conventionally pedestrian tracking such as MOT20~\cite{MOT20}, open-world objects show high dynamicity on shape and motion, which poses significant challenges for motion-based trackers.} %
  \label{fig:ov-motion-example}
\end{figure}

\begin{figure}[t]
  \centering
\includegraphics[width=1\linewidth]{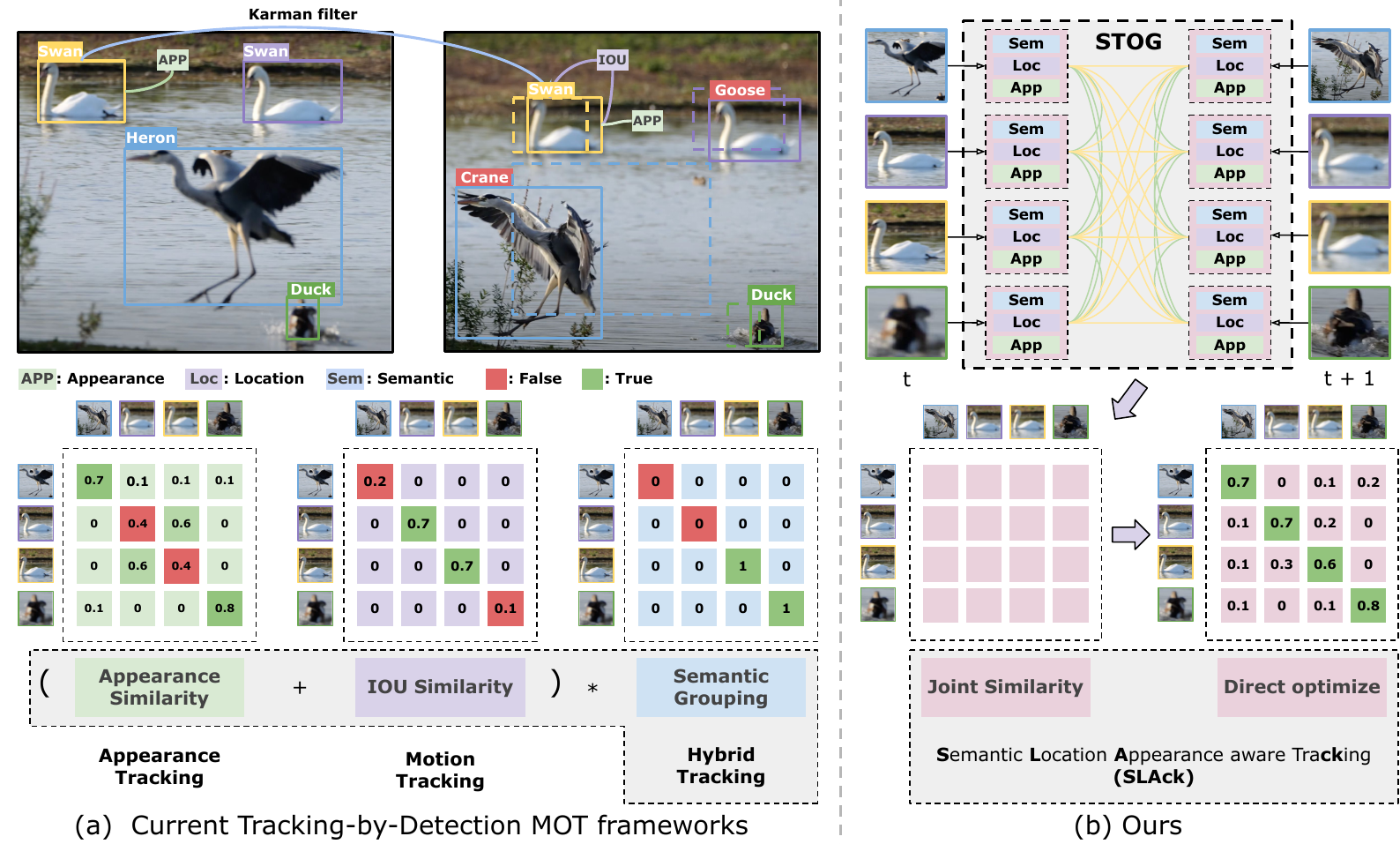}
  \caption{(a) Current MOT methods mainly use motion or appearance cues for tracking. 
  Motion-based trackers typically assume linear motion, employing Kalman Filters, which do not account for the complex, non-linear movements observed in various classes, leading to tracking failures. 
  Appearance-based tracking, which overlooks location information, may confuse targets with similar looks.   Semantic cues are either ignored or used to group instances within the same class as a gating function, which usually leads to errors due to poor classification accuracy in the open world.
  Hybrid trackers combine all three cues—semantic, location, and appearance in a heuristic way which suffers the high dynamicity of the open world.
  (b) Our method jointly fuses the semantic, location, and appearance cues in the early matching stage and yields a single joint similarity matrix that can be directly optimized and trained end-to-end for association.} 
  \label{fig-teaser}
\end{figure}

Multiple Object Tracking (MOT) has traditionally been confined to a limited vocabulary, focusing on categories like pedestrians and vehicles~\cite{MOT17, MOT20, KITTI, nuScenes, Waymo}. The dawn of open-vocabulary tracking ~\cite{OVTrack} expands the horizon to a wider array of classes but also amplifies the challenges, given the diverse appearances, behaviors, and motion patterns across different object classes. Despite the challenges, developing such tracking systems is of great significance. Real-world applications, ranging from autonomous driving to augmented reality, necessitate trackers that transcend these vocabulary constraints for broader, more versatile functionality.

Current best-performing MOT methods for large-vocabulary tracking are based on pure appearances-based matching~\cite{TETer, OVTrack, masa}. Motion-based MOT faces significant challenges since the prevailing motion-based MOT methods lean heavily on the Kalman Filter (KF)~\cite{SORT, DeepSORT, bytetrack, OC-SORT, StrongSORT}. KF-based trackers rely on the linear movement assumption, which is valid in pedestrian or vehicle-centric datasets like MOT Challenges~\cite{MOT17, MOT20}. Unfortunately, this assumption falters in complex open-vocabulary scenarios, characterized by non-linear object movements of varying object classes and motion patterns, as shown by Fig.~\ref{fig:ov-motion-example}.

Despite these challenges, motion and location prior can still provide important cues for robust MOT. Our approach diverges from explicit KF-based motion and, instead, proposes an implicit incorporation of location priors within the association learning process. By projecting location and shape information into the feature space and constructing an implicit spatial and temporal object graph through multi-head attention, our model learns implicit motion priors directly from data. This implicit modeling allows us to learn complex scene-level object structures spatially and enables us to capture both linear and non-linear motion temporally.

The synergy between semantics and motion cues is evident, as motion patterns often correlate with object categories. For instance, if the model learns the motion patterns of horses during training, it can directly use such knowledge to transfer to its semantic similar classes such as zebra during novel class tracking.
This observation fuels our argument for a joint modeling approach, where semantic information enriches the location prior, leading to superior performance in novel class tracking, even without the reliance on appearance cues. 

Appearance features are widely recognized as essential for accurate tracking~\cite{QDTrack, OVTrack}. In our unified framework, these features are seamlessly integrated into the matching process by inclusion in the fused embedding. This integration strategy contrasts sharply with conventional methods that often resort to the heuristic-based fusion at later stages as shown in Fig.~\ref{fig-teaser}.

Our integrated model, named \textbf{SLAck}, encapsulates \textbf{S}emantics, \textbf{L}ocations, and \textbf{A}ppearances within a unified association framework for open-vocabulary tra\textbf{ck}ing. 
Utilizing pre-trained detectors, SLAck extracts a comprehensive set of descriptors and employs an attention-based spatial and temporal object graph (STOG) to facilitate information exchange, both spatially within frames and temporally across frames. Similar processes have been shown useful in modeling spatial object-level relations for object detection~\cite{hu2018relation} or computing correspondence between points~\cite{supergule}. Our motivation is to exchange objects' semantics, appearance, and location, and exploit synergies among these aspects for better novel object tracking.
This process not only enhances the awareness of relative object positions but also aligns motion and appearance patterns with semantic information, resulting in more robust temporal object associations.

 Extensive experiments on large-vocabulary tracking benchmarks affirm the efficacy of our approach, with notable improvements in tracking performance in open-vocabulary settings, especially for novel objects.
 Our results indicate that motion patterns and semantics, when learned implicitly, can significantly enhance tracking performance.
 With the additional integration of appearance cues, our method achieves a substantial boost in association accuracy, confirming the value of our contributions: the early integration of semantic information with appearance and motion-based tracking, the learning of intrinsic cue fusion instead of external heuristics, and the establishment of an implicit spatial-temporal object graph that demonstrates significant generalization in novel class tracking.

\section{Related Work}

\noindent\textbf{Open-Vocabulary MOT}
The advent of the TAO dataset~\cite{TAO}—the first to offer a large vocabulary for tracking with over 800 classes—has spurred research into the open-world MOT challenges~\cite{nettrack, masa,TETer,TAO-OW, OVTrack}. TETA~\cite{TETer} highlights the inadequacy of conventional MOT metrics in large-vocabulary scenarios and introduces a novel evaluation metric. TAO-OW~\cite{TAO-OW} introduces a new setting assessing open-world tracking on TAO. This setting, however, focuses only on recall and is class-agnostic, failing to accurately measure tracker precision and semantic categorization. OVTrack~\cite{OVTrack} proposes an open-vocabulary MOT task that evaluates both precision and recall as well as classification. MASA~\cite{masa} learns universal appearance models for MOT in the open world from vast unlabelled images. Current best-performing open-vocabulary MOT methods are most appearance-based and overlook semantic and location information. We posit that incorporating them can significantly improve generalization in novel class tracking.

\noindent\textbf{MOT with Different Cues}
Motion cues are pivotal in MOT, with methods like SORT~\cite{SORT}, OC-SORT~\cite{OC-SORT}, and ByteTrack~\cite{bytetrack} utilizing the Kalman Filter (KF) for trajectory prediction. Extensions like AM-SORT~\cite{AMSORT} use transformers to adjust KF parameters during occlusions. However, KF-based methods struggle in open-vocabulary scenarios with rapid movement and deformation.

Appearance cues are commonly used in MOT. ~\cite{QDTrack, IDOL, darth, cooler } use discriminative instance appearance embeddings, while TETer~\cite{TETer} and OVTrack~\cite{OVTrack} leverage semantic appearances and Stable Diffusion~\cite{stablediffusion} for association learning. These methods often require extensive annotations and can overfit to base classes, reducing performance on novel classes.

Hybrid approaches combine appearance and motion cues, often through heuristic-based fusion at later stages. DeepSORT~\cite{DeepSORT} enhances SORT with an appearance model for occlusion handling. Other methods ~\cite{JDE, FairMOT, bytetrack, segu2024walker} use dual branches for KF-based IoU and appearance embeddings, combining them heuristically. End-to-end transformer approaches~\cite{MOTR, TransTrack, TrackFormer, segu2024samba} treat instances as queries but face challenges with incomplete open-world video annotations. Offline methods like NeuralSolver~\cite{NeuralSolver} and SUSHI~\cite{sushitrack} model trajectories with future information but require comprehensive annotations and are challenging for real-world deployments, such as robotics.

\section{Method}
We first review popular MOT methods that utilize semantic, motion, appearance, and hybrid cues and investigate their failures when directly employed in open-vocabulary tracking. Then, we introduce our approach \textbf{SLAck} for addressing those issues.

\subsection{Preliminaries: Various cues for MOT}

\begin{figure}[t]
  \centering
\includegraphics[width=1\linewidth]{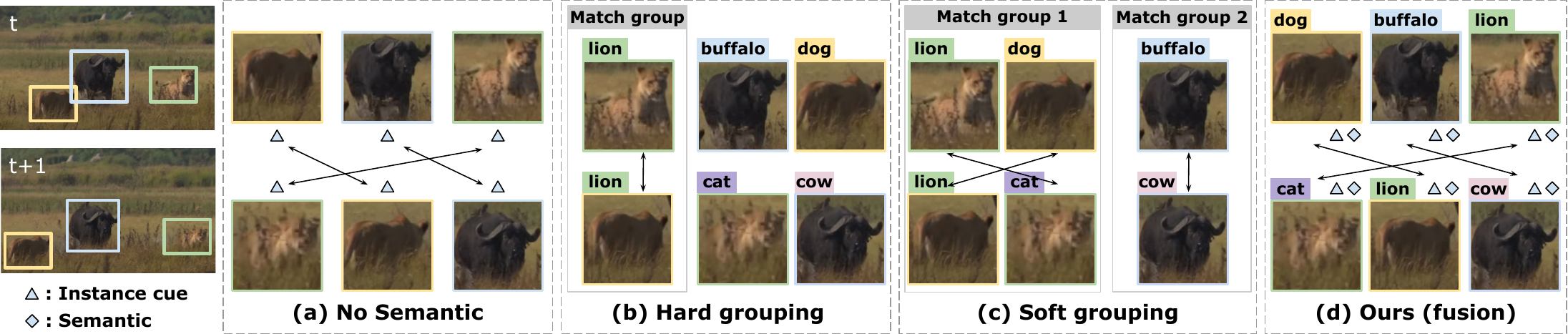}
  \caption{Comparison of different strategies on utilizing semantic cues in open-vocabulary MOT. Instead of leveraging hard grouping or soft grouping~\cite{TETer}, our method integrates semantic cues at the early stage to improve the association accuracy.} %
  \label{fig-usage-semantic}
\end{figure} 

\noindent{}{\textbf{Semantic cues}}
Fig.~\ref{fig-usage-semantic} summarizes different approaches to utilizing semantics in previous MOT literature.
Semantic cues often play a minor role in multi-class MOT, typically utilized as a hard grouping where trackers associate objects of the same class based on detector predictions~\cite{bytetrack, QDTrack}. This approach is effective for simple tasks like tracking humans and vehicles in datasets like KITTI~\cite{KITTI} and nuScenes~\cite{nuScenes}. However, in open-vocabulary tracking, where classification is unreliable, this strategy falls short, as illustrated by Fig.~\ref{fig-teaser}. Relying on such uncertain classifications compromises tracking performance.
TETer~\cite{TETer} proposes using contrastive class exemplar encodings for semantic comparison in feature space, shifting from hard to more dependable soft grouping. Yet, this method still relegates semantic information to a late-stage heuristic-based association. 
In contrast, we advocate for the early integration of semantic cues into the association process, harnessing their informative potential to enhance learning and association accuracy.

\noindent{}{\textbf{Motion cues}}
Most motion-based MOT methods, including ~\cite{SORT, bytetrack, OC-SORT, StrongSORT}, rely on the Kalman Filter (KF) under a linear-motion assumption. However, the dynamic nature of open environments, characterized by changing camera angles, rapid object movements, and diverse motion patterns across classes, challenges the efficacy of linear motion models. This complexity is evidenced both qualitatively, in Fig.~\ref{fig:ov-motion-example}, and quantitatively, through the poor performance of SORT-based trackers indicated in Table~\ref{tab:ovtrack-tao}.
Despite these challenges, object motion remains a valuable cue for tracking, given that spatial structures and proximities are generally consistent, even with non-linear motions. 

In response, we introduce an approach that leverages implicit motion modeling by establishing spatial and temporal relationships between objects. Specifically, we map the location and shape of each object into feature spaces, enabling intra-frame and inter-frame interactions through attention mechanisms. This process facilitates information exchange among objects about their locations, enhancing motion representation without relying on explicit linear assumptions.

\noindent{}{\textbf{Appearance cues}}
Appearance-based methods, including ~\cite{TETer, OVTrack, masa}, predominate in open-vocabulary tracking by utilizing appearance embeddings. These embeddings, derived from an added head to detectors and trained on static images or video pairs via contrastive learning, are crucial for association in diverse tracking scenarios. Yet, reliance on appearance alone introduces challenges like occlusion sensitivity and demands extensive data for learning robust matches, often resulting in overfitting to base categories.

We integrate appearance with semantic information early in the feature-matching process, leveraging the high-level context from semantics while allowing the appearance head to focus on lower-level details which enables us to learn more generalized features across various conditions.

\noindent{}{\textbf{Hybrid cues}}
Methods like JDE, FairMOT, DeepSORT, and ByteTrack~\cite{JDE, FairMOT, DeepSORT, bytetrack} utilize both appearance and motion cues, with motion modeled via the Kalman Filter and appearance derived from either a dedicated re-id network or jointly with detection features. Fusion of these cues occurs at the final matching step, where a spatial proximity matrix (from IoU) and an appearance similarity matrix (via dot product or cosine similarity) are combined through heuristics for Hungarian matching.

Our approach deviates by integrating all valuable information early on, culminating in a singular matching matrix. This early fusion sidesteps the heuristic complexity and enhances generalization, especially for novel classes.

\subsection{Method Overview}
\begin{figure}[t]
  \centering
\includegraphics[width=1\linewidth]{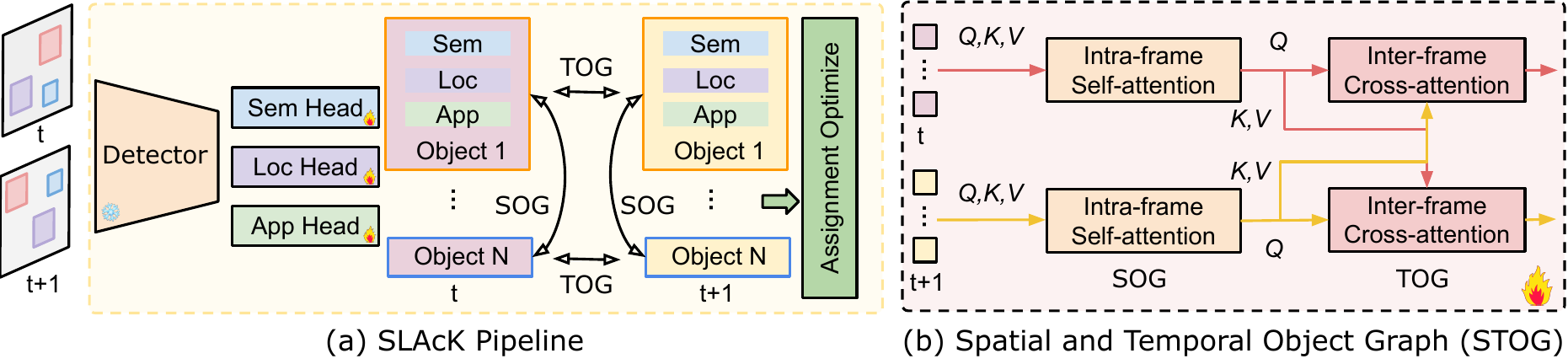}
  \caption{Overall pipeline of SLAck. SLAck first extracts semantic, location, and appearance cues from pre-trained detectors, and then constructs the Spatial-Temporal Object Graph to fuse these cues and model the object dynamics for tracking.} %
  \label{fig-pipeline}
\end{figure}
Our method builds upon pre-trained open or large vocabulary detectors and extends them for tracking.
We extract all information directly from the detectors such as semantics, location, and appearance. We then combine those cues with a spatial-temporal object graph to reason the association assignment. 
Our process is straightforward and end-to-end, without needing any extra heuristics to mix different cues.
The model simply outputs an assignment matrix using the differential Sinkhorn-Knopp algorithm following ~\cite{supergule}. 
Also, to deal with incomplete annotations in the TAO dataset, we directly use both predicted detection boxes and TAO's sparse ground truth as input for the association learning. We call this strategy Detection Aware Training. We show the overall pipeline in the Fig.~\ref{fig-pipeline}.

\subsection{Extract Semantic, Location, and Appearance Cues }
We extract semantic, location, and appearance cues from pre-trained detectors. 
Recently, many strong object detectors~\cite{VILD, DetPro, Detic, groundingdino, ye2023cascade} have emerged for universal object detection. 
To ensure fairness, we build our trackers on the same detectors used in ~\cite{OVTrack, TETer} for open and large vocabulary tracking. 
Importantly, we freeze all detector components during our association process to maintain the original strong open-vocabulary detection capabilities.
We describe the semantic, location, and appearance heads in detail below.

\noindent{}{\textbf{Semantic Head}}
For open-vocabulary tracking, we need to configure the classes we are interested in without re-training. We use the same detector as OVTrack ~\cite{OVTrack}. Directly using a CLIP~\cite{CLIP} encoder for the semantic cues incurs high inference costs. Therefore, the semantic cues come from an adapted RCNN classification head which distillates the CLIP text encoder following~\cite{VILD}. Based on the output embedding of this CLIP-aligned classification head, we add a five-layer MLP to project the semantic feature to get the final semantic embedding $E_\text{sem}$. For the close-set setting, we use TETer's detectors~\cite{TETer} and use their CEM encoding as the input for our semantic head.

\noindent{}{\textbf{{Location Head}}}
The location head takes the output of bounding boxes head from the detector and projects them into feature space. 
Then, the bounding box coordinates are normalized relative to the image dimensions to ensure scale invariance.
Given a set of bounding boxes with coordinates \([x_{\text{min}}, y_{\text{min}}, x_{\text{max}}, \\ y_{\text{max}}]\) for an image of dimensions \([H, W]\), the normalization process involves scaling and translating the coordinates relative to the image center and size. The image center, $(C_x, C_y)$, is computed as $C_x = \frac{W}{2}$, $C_y = \frac{H}{2}$. The scaling factor is determined as $70\%$ of the maximum image dimension, i.e., $scaling = 0.7 \times \max(H, W)$. 
Using this scaling factor, the normalized bounding box coordinates are computed as:
\begin{equation}
(x_{\text{min}}^{\prime}, y_{\text{min}}^{\prime}, w^{\prime}, h^{\prime}) = \left(\frac{x_{\text{min}} - \frac{W}{2}}{s}, \frac{y_{\text{min}} - \frac{H}{2}}{s}, \frac{x_{\text{max}} - x_{\text{min}}}{s}, \frac{y_{\text{max}} - y_{\text{min}}}{s}\right)
\end{equation}
This normalization step ensures that the spatial location of objects within the image is preserved relative to the image dimensions, facilitating scale-invariant location features. 
We then feed the normalized coordinates into the location head to get a location embedding $E_\text{loc}$. For the close-set setting, we include corresponding confidences $c$ with the box coordinates.

\noindent{}{\textbf{Appearance Head}}
The appearance head takes the RoI features embeddings as input and outputs appearance embeddings that are optimized for the association. The appearance head is a simple four-layer convolution with one additional MLP. The RoI embeddings are transformed by the appearance head and get an appearance embedding $E_\text{app}$.

\subsection{Spatial-Temporal Object Graph (STOG)}
Spatial-Temporal Object Graph (STOG) accomplishes this by leveraging a combination of intra-frame self-attention and inter-frame cross-attention mechanisms following ~\cite{supergule} across different objects, encoding rich semantic, location, and appearance patterns that are essential for understanding object dynamics during open-world tracking.

\noindent{}{\textbf{Feature Fusion in STOG}}
Before STOG, each object within a frame is represented by a set of distinct yet complementary features: semantic embeddings ($E_\text{sem}$), location ($E_\text{loc}$), and appearance ($E_\text{app}$). These embeddings capture various aspects of an object, such as its visual characteristics, spatial position, and class information, which are crucial for accurate tracking.
The STOG model initiates its process by fusing the appearance, location, and semantic features of each object into a unified representation. The fusion is conducted through a simple yet effective summation operation:
\begin{equation}
    E^i_{\text{fused}} = E^i_{\text{app}} + E^i_{\text{loc}} + E^i_{\text{sem}},
\end{equation}
where $E^i_{\text{fused}}$ represents the fused embedding for the $i$-th object in the frame. This operation ensures that the information from all three modalities is integrated, allowing the STOG to process each object with a comprehensive understanding of its appearance, spatial context, and semantic meaning.

The fused embeddings serve as the input to the STOG model, where they undergo further processing through the intra-frame self-attention and inter-frame cross-attention mechanisms. The self-attention mechanism within each frame allows the model to refine its understanding of the objects based on their relative positions and visual similarities, while the cross-attention mechanism across frames captures the temporal evolution of objects, facilitating robust tracking even in complex dynamic scenes.

\noindent{}{\textbf{Intra-Frame Self-Attention}}
The intra-frame self-attention mechanism independently processes objects within both the key and reference frames to analyze their spatial relationships and interactions. For objects in the keyframe (\(K\)) and reference frame (\(R\)), the self-attention (SA) operations are defined as follows:
{\small
\begin{equation}
    \text{SA}_{K}(Q_{K}, K_{K}, V_{K}) = \sigma \left(\frac{Q_{K}K_{K}^T}{\sqrt{d}}\right)V_{K},
    \text{SA}_{R}(Q_{R}, K_{R}, V_{R}) = \sigma \left(\frac{Q_{R}K_{R}^T}{\sqrt{d}}\right)V_{R},
\end{equation}}
\noindent where $\sigma$ denotes the Softmax operation, \(Q_{K}, K_{K}, V_{K}\) and \(Q_{R}, K_{R}, V_{R}\) denote the query, key, and value vectors for objects within the key and reference frames, respectively. \(d\) represents the dimensionality of the vectors. This step enhances the model's understanding of the complex intra-frame object arrangements and interactions by focusing on the most contextually relevant features of each object.

\noindent{}{\textbf{Inter-Frame Cross-Attention}}
After processing intra-frame relationships, the model applies inter-frame cross-attention (CA) to both the key and reference frames independently, aiming to align and update object features across frames. This captures the temporal dependencies essential for tracking objects. The cross-attention operations for objects transitioning from the key to the reference frame and vice versa are formalized as:

{\small
\begin{equation}
    \text{CA}_{K\rightarrow R}(Q_{K}, K_{R}, V_{R}) = \sigma\left(\frac{Q_{K}K_{R}^T}{\sqrt{d}}\right)V_{R},
    \text{CA}_{R\rightarrow K}(Q_{R}, K_{K}, V_{K}) = \sigma\left(\frac{Q_{R}K_{K}^T}{\sqrt{d}}\right)V_{K},
\end{equation}}

where \(Q_{K}\) and \(Q_{R}\) are the query vectors derived from objects in the key and reference frames, respectively, while \(K_{R}, V_{R}\) and \(K_{K}, V_{K}\) are the key and value vectors from the opposite frames. 
$\sigma$ denotes the Softmax operation.
This step is crucial for tracking objects over time, enabling the model to emphasize features that are most relevant for understanding the temporal evolution of each object.
By first applying self-attention within each frame and then cross-attention across frames, the STOG model effectively captures both spatial and temporal object dynamics, enhancing its ability to track multiple objects across frames in a video.

\subsection{Association Loss}

Given a pair of frames, the objective is to compute a loss that encourages correct matching between objects across these frames. This process involves generating a target match matrix and then applying the Sinkhorn algorithm to compute a differentiable approximation of the optimal transport problem. The loss is calculated as follows:

\noindent{}{\textbf{Target Match Matrix Computation}}
For each pair of key and reference frames, we construct a binary target match matrix, $\mathbf{T}$, to represent matches between objects across frames. Each entry in $\mathbf{T}$, denoted as $T_{ij}$, is set to 1 if the $i$-th object in the key frame matches the $j$-th object in the reference frame, and 0 otherwise. This method relies on ground truth match indices to determine the correspondence between objects. To accommodate objects that do not find a match in the counterpart frame, indicative of objects disappearing or newly appearing, we use a special 'dustbin' class. This adjustment expands the matrix to $\mathbf{T}^\prime$ with dimensions $(M+1) \times (N+1)$, where $M$ and $N$ signify the count of objects in the key and reference frames, respectively.

\noindent{}{\textbf{Training Loss}}
The whole open-vocabulary tracking framework can be trained in an end-to-end manner using the Sinkhorn algorithm, which provides a differentiable solution to the optimal transport problem. Given a score matrix $\mathbf{S}$ from the model, representing the predicted affinities between objects in the key and reference frames, and the augmented target match matrix $\mathbf{T}^\prime$, the Sinkhorn loss $\mathcal{L}_{\text{Sinkhorn}}$ is computed as follows:

\begin{equation}
\mathcal{L}_{\text{Sinkhorn}} = -\sum_{i,j} T_{ij}^\prime \log(S_{ij}^\prime),
\end{equation}
where $S_{ij}^\prime$ is the softmax-normalized score matrix after applying the Sinkhorn iterations, akin to a probabilistic matching.
The final loss $\mathcal{L}$ for a batch of frames is the average of $\mathcal{L}_{\text{Sinkhorn}}$ over all frame pairs, encouraging the model to predict matches that align with the ground truth correspondences encoded in $\mathbf{T}^\prime$.

\subsection{Detection Aware Training (DAT)}
Open-vocabulary tracking faces a significant challenge due to limited training data, with TAO~\cite{TAO} being the sole large-vocabulary MOT dataset, characterized by its incomplete annotations. This issue arises from the impracticality of exhaustively annotating every object in video sequences, compounded by the expense of video annotation and the complexity of class definitions, which may vary in granularity.
To address the incomplete annotation problem without compromising the detection capabilities of our pre-trained open-vocabulary detector, we freeze the detector's weights during training on TAO videos. This ensures the detector's performance remains unchanged. To adapt to the sparse annotations, we employ a strategy where the detector first infers bounding boxes on training videos, maintaining consistency in input data during both training and testing phases. We only compute the association loss when there is a matching between these predicted boxes and available ground truth, disregarding unmatched pairs. This method proves highly effective for end-to-end training in MOT tasks, allowing for significant performance gains by closely replicating test conditions during the training process.

\section{Experiment}

\subsection{Evaluation Metrics}
Following ~\cite{OVTrack, GLEE}, we use \textbf{TETA} metrics ~\cite{TETer} for the large-scale multiple-class multiple object tracking. TETA disentangles MOT evaluation into three sub-factors: Localization  (LocA), Association (AssocA) and Classification (ClsA). 
Additionally, TETA supports evaluation with incomplete annotations, which gives more accurate tracking measurement for the open-vocabulary MOT~\cite{OVTrack}.

\subsection{Benchmarks}
We evaluate our methods using the TAO dataset~\cite{TAO}, the only large-scale, open-vocabulary MOT dataset. With over 800 categories, TAO stands as the most comprehensive MOT dataset regarding class diversity to date. It encompasses 500, 988, and 1,419 40-second annotated videos across train, validation, and test sets, respectively. TAO offers various benchmarks, highlighting distinct characteristics and requirements. We only use the TAO training set for training.

\noindent{}{\textbf{Open-Vocabulary MOT~\cite{OVTrack}}}
This benchmark challenges trackers to generalize across novel classes without training on their annotations. TAO's adherence to the LVIS~\cite{LVIS} taxonomy means the split between base and novel classes aligns with established open-vocabulary detection protocols~\cite{VILD}, designating frequent and common classes as the base and rare ones as the novel. This setup reflects real-world conditions where trackers encounter seldom-seen categories, testing their adaptability and stability in tracking uncommon objects. 

\noindent{}{\textbf{TAO TETA~\cite{TETer}}}
TAO TETA benchmark is a closed-set MOT challenge, permitting training on all class annotations within TAO. It focuses on association quality, rewarding trackers that generate precise trajectories without overlaps.
\subsection{Implementation Details}
Our model, \ourmodel-OV, is constructed atop pre-trained open and large vocabulary detectors. For open-vocabulary MOT, we employ the same Faster R-CNN detector as OVTrack~\cite{OVTrack}, featuring a ResNet-50~\cite{ResNet} backbone trained on LVIS base classes. 
For the close-set TAO TETA benchmark, we utilize Faster R-CNN detectors with a class exemplar head, akin to TETer~\cite{TETer}. The variants \ourmodel-T and \ourmodel-L deploy Swin-Tiny and Swin-Large backbones~\cite{swin}, respectively. 
Training images are randomly resized, keeping the aspect ratio, with the shorter side between 640 and 800 pixels. Pairs of adjacent frames, within a maximum interval of 3 seconds, are selected for training. Further details are provided in the appendix.

\subsection{Ablation Study}
We provide comprehensive ablation experiments to validate the effectiveness of our proposed models and training strategy. Ablation tables report novel tracking performance on the open-vocabulary MOT benchmark~\cite{OVTrack} if not specified.

\noindent\textbf{Effectiveness of semantic-aware matching}
\begin{table}[t]
    \caption{\textbf{Effectiveness of semantic-aware matching.} We examine the effectiveness of the semantic cues in novel class tracking. We focus on AssocA, which evaluates the association's performance. We observe consistent improvement when we add the semantic cues into the early association process. }%
    \centering%
    \resizebox{0.8\linewidth}{!}{\resizebox{1\linewidth}{!}{
\begin{tabular}{l|ccc|cccc}
\toprule
\multicolumn{1}{c|}{Method} & \multicolumn{3}{c|}{Cues} & \multicolumn{4}{c}{Novel}\\ 
\midrule
TAO-val & Semantic & Location & Appearance & TETA & \textcolor{gray!50}{LocA} & AssocA & \textcolor{gray!50}{ClsA}  \\ \midrule
\underline{OC-SORT} & - & \checkmark & - & 23.7 & \textcolor{gray!50}{49.6} & 20.4 & \textcolor{gray!50}{1.1} \\
Lck (Motion-only)  & - & \checkmark & - & 27.8 & \textcolor{gray!50}{54.3} & 28.3 & \textcolor{gray!50}{0.9} \\
SLck (Semantic-aware Motion) & \checkmark & \checkmark & - & 30.5 (\textcolor{green}{+2.2}) & \textcolor{gray!50}{55.1} & 35.4 (\textcolor{green}{+7.1}) & \textcolor{gray!50}{1.0} \\ \midrule
\underline{OVTrack} & - & - & \checkmark & 27.8	 & \textcolor{gray!50}{48.3} & 33.6 & \textcolor{gray!50}{1.5} \\ 
Ack (Appearance-only) & - & - & \checkmark & 29.4 & \textcolor{gray!50}{54.3} & 32.7 & \textcolor{gray!50}{1.1} \\
SAck (Semantic-aware Appearance) & \checkmark & - & \checkmark & 29.7 (\textcolor{green}{+0.3}) & \textcolor{gray!50}{53.2} & 35.1 (\textcolor{green}{+2.4}) & \textcolor{gray!50}{0.9} \\ \midrule
LAck (Motion-Appearance)& - & \checkmark & \checkmark & 30.8 & \textcolor{gray!50}{54.6} & 36.4 & \textcolor{gray!50}{1.4} \\
\textbf{SLAcK (Full model)} & \checkmark & \checkmark & \checkmark & \textbf{31.1} (\textcolor{green}{+0.3}) & \textcolor{gray!50}{55.0} & \textbf{37.8} (\textcolor{green}{+1.4}) & \textcolor{gray!50}{1.3} \\ 
\bottomrule
\end{tabular}
}

}%
    \label{tab:semantic-matching}%
\end{table}
Previously, semantic cues are largely ignored by MOT methods. The association is done by either using motion or appearance cues. We validate the importance of semantic information for large-scale open-vocabulary tracking, especially in novel classes. We also add the pure motion-based OC-SORT~\cite{OC-SORT} and pure appearance-based OVTrack~\cite{OVTrack} as baselines for easier comparison. Table~\ref{tab:semantic-matching} presents the results. Our model with location 
cues only (Lck) simulates the motion-based trackers whose association is done purely based on location cues. Adding semantics improves the association performance significantly by an increase of +7.1 on AssocA. It shows that modelling the semantics together with location or motion cues can effectively enhance the novel class tracking performance. With our semantic-aware motion tracker (SLck), we already outperform the appearance-based state-of-the-art method OVTrack~\cite{OVTrack} by +1.8 on AssocA. This further demonstrates our assumption that semantic and motion patterns are highly related. The motion patterns learned on base classes can effectively transfer to the unseen but semantically similar novel classes, to improve the generalization of trackers. Adding semantics cues into appearance-base matching (SAck) also increases the AssocA by +2.4 compared to the appearance-only method (Ack). Finally, adding the semantic cues can also help the hybrid tracker (LAck) which includes both motion and appearance cues by further +1.4 improvement on AssocA. This demonstrates the importance of incorporating semantic cues into associations. 

\noindent\textbf{Effectiveness on DAT}
We evaluate the effectiveness of our proposed detection-aware training (DAT) strategy. Table~\ref{tab:dat} shows that training with DAT significantly improves the association performance by (+13.7 AssocA) compared to the training directly with the sparse and incomplete GroundTruth. DAT exploits the sparse supervision and, in the meantime, simulates the training with inference by using all predicted bounding boxes for training which significantly eliminates the gap between training and inference. 

\noindent\textbf{Comparison of different ways of utilizing semantic cues} We benchmark against existing methods for utilizing semantic cues on the TAO TETA benchmark. We conduct these experiments based on TETer-T. Our investigation reveals that traditional hard grouping~\cite{QDTrack, bytetrack}, which associates objects sharing identical predicted class labels, underperforms significantly, showing a -4.6 decrease in AssocA. This drop is attributed to the instability of class predictions in large vocabularies, where hard classification constraints can hurt the overall tracking accuracy. Alternatively, soft grouping, which clusters semantically similar objects based on feature similarity, offers varied performance improvements depending on the features employed. Utilizing CEM encoding, a contrastive semantic feature trained on LVIS, soft grouping yields a +1.5 increase in AssocA over the no-semantic baseline. However, using BERT encoding hurt the performance. We add our semantic-aware Matching (Ours-SAck) to the TETer-T by integrating CEM encoding and appearance features directly within the STOG association process, eliminating late-stage grouping or gating mechanisms. This approach results in a substantial improvement, outperforming the no-semantic baseline by +2.8 and the CEM soft grouping by +1.3 in AssocA. This gain underscores the efficiency of our semantic-aware matching strategy in leveraging semantic information for large-vocabulary tracking.

\noindent\textbf{Are semantic cues alone enough for tracking?}
As shown in Table~\ref{tab:stog}, we also check whether the semantic cues alone contain sufficient appearance information that is distinguishable enough for tracking. Compared to the appearance alone (Ack-STOG), the association accuracy drops significantly -4.4 AssocA using semantic cues (Sck-STOG) alone.  It suggests that although semantic cues contain instance appearance, however, they are not as distinguishable as the task-specific appearance features. 

\noindent\textbf{Effectiveness of STOG}
We assess the impact of spatial (SOG) and temporal (TOG) object graphs on trackers using different cues: semantic (Sck), location (Lck), appearance (Ack), and combination (SLAck). Table~\ref{tab:stog} compares trackers with and without SOG and TOG. Ind means without STOG. It uses simple MLP and addition to obtain the final tracking features. The results in Table~\ref{tab:stog} indicate TOG is important for semantic (+2.4 in AssocA) or location cues (+4.1 in AssocA). TOG provides insight into which objects are present in subsequent frames, thereby refining the features for better associations. For example, location-based trackers can anticipate potential object movements, while semantic-based trackers can refine the semantic encodings based on temporal information.
Appearance-based trackers benefit more from SOG (+0.9 in AssocA compared to TOG's +0.3). SOG facilitates awareness of co-located objects' appearances, encouraging feature differentiation to improve association accuracy. When all three cues are integrated with both SOG and TOG (STOG), we observe a substantial boost in association performance (+3.7 in AssocA). This underscores the synergistic effect of STOG with multiple tracking cues, significantly enhancing association accuracy.

\begin{table}[!t]
			\centering%
		\begin{minipage}[t]{0.44\linewidth}

  \caption{\textbf{Training with incomplete annotated GT vs Ours DAT.}}%
    \centering%
    \resizebox{\linewidth}{!}{\begin{tabular}{l|cccc}
\toprule
Method & TETA & LocA & AssocA & ClsA \\
\midrule
SLAck with sparse GT & 24.6 & 47.1 & 24.1 & \textbf{2.5} \\
SLAck with DAT & \textbf{31.1} & \textbf{54.3} & \textbf{37.8} (\textcolor{green}{+13.7}) & 1.3 \\
\bottomrule
\end{tabular}
}%
    \label{tab:dat}%
		\end{minipage}
		\begin{minipage}[t]{0.53\linewidth}
			\caption{\textbf{Comparison of different ways of integrating semantic cues for association.} The comparison is done on the TAO TETA benchmark and we only focus on the AssocA. The study is based on TETer-T.}%
    \centering%
    \resizebox{1.0\linewidth}{!}{\begin{tabular}{l|c|c|c|c|c}
\toprule
 & No & Hard Grouping & Soft Grouping & Soft Grouping  & Semantic-aware Matching \\
 & semantic & (Class) & (BERT Embed) & (CEM Embed) & (Ours-SAck) \\
\midrule
AssocA & 35.2 & 30.6 & 31.2 & 36.7 & \textbf{38.0} \\
\bottomrule
\end{tabular}
}%
    \label{tab:different_semantic}%
    \end{minipage}
	\end{table}

\begin{table}[t]
\caption{\textbf{Open-vocabulary MOT comparison.} We compare our method on open-vocabulary MOT benchmark~\cite{OVTrack}. We indicate the classes and data the methods trained on. All methods use ResNet50 as the backbone. $^{\dagger}$ represents use the same detector.}
     \centering%
     \resizebox{.7\linewidth}{!}{\resizebox{1\linewidth}{!}{
\begin{tabular}{l|cc|cccc|cccc}
\toprule
\multicolumn{1}{c|}{Method}  & \multicolumn{2}{c|}{Classes} & \multicolumn{4}{c|}{Novel}   & \multicolumn{4}{c}{Base}\\ \midrule
\textbf{Validation set} & Novel & Base & TETA & LocA & AssocA & ClsA & TETA & LocA & AssocA & ClsA  \\ \midrule
QDTrack~\cite{QDTrack} & \checkmark & \checkmark & 22.5 &42.7 &24.4 & 0.4 & 27.1& 45.6 &24.7 &11.0   \\
TETer~\cite{TETer} & \checkmark & \checkmark & 25.7 &45.9 &31.1 & 0.2 & 30.3 &47.4 &31.6 &12.1  \\ 
DeepSORT (ViLD)~\cite{DeepSORT} & - & \checkmark & 21.1 & 46.4 & 14.7 & \textbf{2.3} & 26.9 & 47.1 & 15.8 & 17.7 \\
Tracktor++ (ViLD)~\cite{Tracktor} & - & \checkmark & 22.7 & 46.7 & 19.3 & 2.2 & 28.3 & 47.4 & 20.5 & 17.0   \\ 
ByteTrack$^{\dagger}$~\cite{bytetrack} & - & \checkmark & 22.0 & 48.2 & 16.6 & 1.0 & 28.2 & 50.4 & 18.1 & 16.0 \\ 
OC-SORT$^{\dagger}$~\cite{OC-SORT} & - & \checkmark & 23.7 & 49.6 & 20.4 & 1.1 & 28.9 & 51.4 & 19.8 & 15.4 \\ 
{OVTrack$^{\dagger}$~\cite{OVTrack}}  & - & \checkmark & {27.8} & {48.8} & {33.6} & {1.5} & {35.5} & {49.3} & {36.9} & \textbf{20.2} \\ 
MASA (R50)$^{\dagger}$~\cite{masa} & - & - & {30.0} & {54.2} & {34.6} & {1.0} & {36.9} & {55.1} & {36.4} & {19.3} \\ 
\textbf{\ourmodel-OV}$^{\dagger}$   & - & \checkmark & \textbf{31.1} & \textbf{54.3} & \textbf{37.8} & {1.3} & \textbf{37.2} & \textbf{55.0} & \textbf{37.6} & {19.1} \\ \midrule
\textbf{Test set} & Novel & Base & TETA & LocA & AssocA & ClsA & TETA & LocA & AssocA & ClsA \\ \midrule
QDTrack~\cite{QDTrack} & \checkmark & \checkmark & 20.2 & 39.7 & 20.9 & 0.2 &25.8& 43.2 &23.5 &10.6 \\
TETer~\cite{TETer} & \checkmark & \checkmark & 21.7 & 39.1 & 25.9 & 0.0 & 29.2 & 44.0 & 30.4 & 10.7 \\ 
DeepSORT (ViLD)~\cite{DeepSORT} & - & \checkmark & 17.2 & 38.4 & 11.6 & 1.7 & 24.5 & 43.8 & 14.6 & 15.2 \\
Tracktor++ (ViLD)~\cite{Tracktor} & - & \checkmark & 18.0 & 39.0 & 13.4 & 1.7 & 26.0 & 44.1 & 19.0 & 14.8 \\ 
{OVTrack$^{\dagger}$~\cite{OVTrack}}  & - & \checkmark & {24.1} & {41.8} & {28.7} & {1.8} & {32.6}& {45.6} & \textbf{35.4} & \textbf{16.9} \\
\textbf{\ourmodel-OV$^{\dagger}$}  & - & \checkmark & \textbf{27.1} & \textbf{49.1} & \textbf{30.0} & \textbf{2.0} & \textbf{34.7} & \textbf{52.5} & {35.6} & {16.1} \\ 
\bottomrule
\end{tabular}
}
}%
    \label{tab:ovtrack-tao}%
\end{table}

\subsection{Comparison to State-of-the-Art}
\noindent\textbf{Open-vocabulary MOT} Table~\ref{tab:ovtrack-tao} presents our evaluation of the validation and test splits for open-vocabulary MOT~\cite{OVTrack}. To maintain comparability, we utilize a ResNet-50 backbone across all methods. QDTrack~\cite{QDTrack} and TETer~\cite{TETer}, serving as closed-set baselines, train using all available classes. Competing methods enhance existing open-vocabulary detectors by incorporating off-the-shelf trackers like ByteTrack,  OC-SORT, and MASA. OVTrack, previously the top performer in open-vocabulary MOT, leverages Stable Diffusion~\cite{stablediffusion} to augment LVIS images for base class training. On val split, our model, \ourmodel-OV, significantly surpasses all existing methods in tracking novel classes, particularly in association accuracy (AssocA), with a +4.2 improvement over OVTrack. This gap underscores OVTrack's limitation as a pure appearance-based tracker.

\noindent{}\textbf{TAO TETA} We also compare our methods on the close-set setting with all existing works in the established TAO TETA benchmark. As shown in Table~\ref{tab:tao-teta}, our ~\ourmodel-L outperforms all existing methods in AssocA, including methods trained with additional tracking annotations such as UNINEXT~\cite{UNINEXT} and GLEE~\cite{GLEE}. Using the same detectors as TETer-T and TETer-L~\cite{TETer}, our models outperform each by +2.2 and +1.9 on the AssocA. 
Moreover, we outperform the second-best GLEE-Plus with the Swin-Large backbone by +0.9 in AssocA. Note that GLEE is a foundation vision model trained with 10 million images. This demonstrates the strong generalization and superior association accuracy of our methods.

\begin{table}[!t]
			\centering%
		\begin{minipage}[t]{0.47\linewidth}

       \caption{\textbf{Ablation on the STOG.} Ind: without STOG, SOG: Spatial Object Graph. TOG: Temporal Object Graph, Sck: using only Semantic, Lck: using only Location. SLAck: our full model. }%
    \centering%
    \resizebox{1\linewidth}{!}{

\begin{tabular}{l|cc|cccc}
\toprule
Method & Spatial & Temporal & TETA & \textcolor{gray!50}{LocA} & AssocA & \textcolor{gray!50}{ClsA} \\
\midrule
Sck - Ind & - & - & 27.6 & \textcolor{gray!50}{52.9} & 28.7 & \textcolor{gray!50}{1.1} \\
Sck - SOG & \checkmark & - & 27.5 & \textcolor{gray!50}{53.3} & 28.1 & \textcolor{gray!50}{1.0} \\
Sck - TOG & - & \checkmark & 28.6 & \textcolor{gray!50}{53.4} & 31.1 & \textcolor{gray!50}{1.1} \\
Sck - STOG & \checkmark & \checkmark & 27.6 & \textcolor{gray!50}{53.3} & 28.4 & \textcolor{gray!50}{1.1} \\ \midrule
Lck - Ind & - & - & 26.8 & \textcolor{gray!50}{54.2} & 24.8 & \textcolor{gray!50}{1.4} \\
Lck - SOG & \checkmark & - & 27.3 & \textcolor{gray!50}{53.9} & 26.6 & \textcolor{gray!50}{1.5} \\
Lck - TOG & - & \checkmark & 28.2 & \textcolor{gray!50}{54.5} & 28.9 & \textcolor{gray!50}{1.2} \\
Lck - STOG & \checkmark & \checkmark & 27.8 & \textcolor{gray!50}{54.3} & 28.3 & \textcolor{gray!50}{0.9} \\ 
\midrule
Ack - Ind & - & - & 28.4 & \textcolor{gray!50}{53.8} & 30.4 & \textcolor{gray!50}{0.8} \\
Ack - SOG & \checkmark & - & 28.8 & \textcolor{gray!50}{54.2} & 31.3 & \textcolor{gray!50}{0.8} \\
Ack - TOG & - & \checkmark & 28.4 & \textcolor{gray!50}{53.3} & 30.7 & \textcolor{gray!50}{1.2} \\
Ack - STOG & \checkmark & \checkmark & 29.4 & \textcolor{gray!50}{54.3} & 32.7 & \textcolor{gray!50}{1.1} \\
\midrule
SLAck - Ind & - & - & 29.6 & \textcolor{gray!50}{53.9} & 34.1 & \textcolor{gray!50}{0.8} \\
SLAck - SOG & \checkmark & - & 30.1 & \textcolor{gray!50}{53.9} & 35.5 & \textcolor{gray!50}{0.8} \\
SLAck - TOG & - & \checkmark & 29.7 & \textcolor{gray!50}{53.5} & 34.7 & \textcolor{gray!50}{0.9} \\
SLAck - STOG & \checkmark & \checkmark & 31.1 & \textcolor{gray!50}{54.3} & 37.8 & \textcolor{gray!50}{1.3} \\
\bottomrule
\end{tabular}
}%
    \label{tab:stog}
		\end{minipage}
		\label{tab:test1}
		\hfill
		\begin{minipage}[t]{0.49\linewidth}
			\caption{\textbf{Comparison on TAO TETA Benchmark}~\cite{TETer}.  Our models achieve better performance compared to SOTA  methods using the same detectors. $^{\dagger}$ indicates using the same object detector as TETer-T~\cite{TETer}.}%
    \centering%
    \resizebox{0.9\linewidth}{!}{\resizebox{1\linewidth}{!}{
  \begin{tabular}{lcccc}
    \specialrule{.1em}{.05em}{.05em} 
    Method & TETA & LocA & AssocA & ClsA \\
    \hline%
    SORT~\cite{SORT} & 24.8 & 48.1 & 14.3 & 12.1 \\
    Tracktor~\cite{Tracktor} & 24.2  & 47.4 & 13.0 & 12.1 \\
    DeepSORT~\cite{DeepSORT} & 26.0  & 48.4 & 17.5 & 12.1 \\
    Tracktor++~\cite{Tracktor} & 28.0  & 49.0 & 22.8 & 12.1 \\
    QDTrack~\cite{QDTrack} & 30.0  & 50.5 & 27.4 & 12.1 \\
    OVTrack ~\cite{OVTrack}  & 34.7 & 49.3 & 36.7 & 18.1 \\ 
    ByteTrack$^{\dagger}$ ~\cite{bytetrack} & 27.6 & 48.3& 20.2 & 14.4\\
    OC-SORT$^{\dagger}$~\cite{OC-SORT} & 28.6 & 49.7& 21.8 & 14.3\\
    MASA(R50)$^{\dagger}$ ~\cite{masa} &34.1 &52.1&35.7&15.0 \\
    MASA(Detic)$^{\dagger}$ ~\cite{masa} &34.7 &51.9& 36.4 & 15.8 \\
    MASA(GroundingDINO)$^{\dagger}$~\cite{masa} &34.9 &51.8& 37.6 & 15.4 \\ \midrule
    TETer-T$^{\dagger}$~\cite{TETer}& 34.6 & 52.1 & {36.7} & 15.0 \\
    \textbf{\ourmodel-T}$^{\dagger}$ &  \textbf{35.5} & \textbf{52.2} & \textbf{38.9} & \textbf{15.6} \\
    TETer-L~\cite{TETer} & 40.1 & 56.3 & {39.9} & 24.1 \\  
    \textbf{\ourmodel-L} &  \textbf{41.1} & \textbf{56.3} & \textbf{41.8} & \textbf{25.1} \\ \midrule
    \underline{\textit{Train with additional tracking labels}}& & &&\\
    \textcolor{gray!50}{AOA~\cite{AOA}} & \textcolor{gray!50}{25.3} & \textcolor{gray!50}{23.4} & \textcolor{gray!50}{30.6} & \textcolor{gray!50}{{21.9}} \\
    \textcolor{gray!50}{UNINEXT(R50)~\cite{UNINEXT}} & \textcolor{gray!50}{31.9} & \textcolor{gray!50}{43.3} & \textcolor{gray!50}{35.5} & \textcolor{gray!50}{17.1} \\
    \textcolor{gray!50}{GLEE-Plus(SwinL) ~\cite{GLEE}} & \textcolor{gray!50}{41.5} & \textcolor{gray!50}{52.9} & \textcolor{gray!50}{40.9} & \textcolor{gray!50}{30.8} \\ 
   \specialrule{.1em}{.05em}{.05em} 
  \end{tabular}}
}%
    \label{tab:tao-teta}
		\end{minipage}%
	\end{table}

\vspace{-0.1in}
\section{Conclusion}
We introduce SLAck, a unified open-vocabulary MOT approach that ingeniously integrates semantic and location cues with appearance features for object association. This framework, underpinned by a dynamic spatial-temporal object graph, adeptly captures the intricate relationships between objects, leveraging both their visual attributes and contextual information. By removing traditional late-stage fusion heuristics and in favor of early integration, the method not only simplifies the computational process but also enhances tracking accuracy, especially for novel object classes. 
The promising results on the large-vocabulary MOT benchmarks underscore the superior generalization capabilities of the proposed method.

\bibliographystyle{splncs04}
\bibliography{egbib}

\begin{thebibliography}{10}
\providecommand{\url}[1]{\texttt{#1}}
\providecommand{\urlprefix}{URL }
\providecommand{\doi}[1]{https://doi.org/#1}

\bibitem{Tracktor}
Bergmann, P., Meinhardt, T., Leal-Taixe, L.: Tracking without bells and whistles. In: ICCV (2019)

\bibitem{SORT}
Bewley, A., Ge, Z., Ott, L., Ramos, F., Upcroft, B.: Simple online and realtime tracking. In: ICIP (2016)

\bibitem{NeuralSolver}
Bras{\'o}, G., Leal-Taix{\'e}, L.: Learning a neural solver for multiple object tracking. In: Proceedings of the IEEE/CVF conference on computer vision and pattern recognition. pp. 6247--6257 (2020)

\bibitem{nuScenes}
Caesar, H., Bankiti, V., Lang, A.H., Vora, S., Liong, V.E., Xu, Q., Krishnan, A., Pan, Y., Baldan, G., Beijbom, O.: nuscenes: A multimodal dataset for autonomous driving. In: Proceedings of the IEEE/CVF conference on computer vision and pattern recognition. pp. 11621--11631 (2020)

\bibitem{OC-SORT}
Cao, J., Pang, J., Weng, X., Khirodkar, R., Kitani, K.: Observation-centric sort: Rethinking sort for robust multi-object tracking. In: Proceedings of the IEEE/CVF Conference on Computer Vision and Pattern Recognition. pp. 9686--9696 (2023)

\bibitem{sushitrack}
Cetintas, O., Bras\'o, G., Leal-Taix\'e, L.: Unifying short and long-term tracking with graph hierarchies. In: Proceedings of the IEEE/CVF Conference on Computer Vision and Pattern Recognition (CVPR). pp. 22877--22887 (June 2023)

\bibitem{TAO}
Dave, A., Khurana, T., Tokmakov, P., Schmid, C., Ramanan, D.: {TAO}: A large-scale benchmark for tracking any object. In: ECCV (2020)

\bibitem{MOT20}
Dendorfer, P., Rezatofighi, H., Milan, A., Shi, J., Cremers, D., Reid, I., Roth, S., Schindler, K., Leal-Taix{\'e}, L.: Mot20: A benchmark for multi object tracking in crowded scenes. arXiv preprint arXiv:2003.09003  (2020)

\bibitem{AOA}
Du, F., Xu, B., Tang, J., Zhang, Y., Wang, F., Li, H.: 1st place solution to eccv-tao-2020: Detect and represent any object for tracking. arXiv preprint arXiv:2101.08040  (2021)

\bibitem{DetPro}
Du, Y., Wei, F., Zhang, Z., Shi, M., Gao, Y., Li, G.: Learning to prompt for open-vocabulary object detection with vision-language model. In: Proceedings of the IEEE/CVF Conference on Computer Vision and Pattern Recognition. pp. 14084--14093 (2022)

\bibitem{StrongSORT}
Du, Y., Zhao, Z., Song, Y., Zhao, Y., Su, F., Gong, T., Meng, H.: Strongsort: Make deepsort great again. IEEE Transactions on Multimedia  (2023)

\bibitem{KITTI}
Geiger, A., Lenz, P., Stiller, C., Urtasun, R.: Vision meets robotics: The kitti dataset. The International Journal of Robotics Research  \textbf{32}(11),  1231--1237 (2013)

\bibitem{VILD}
Gu, X., Lin, T.Y., Kuo, W., Cui, Y.: Open-vocabulary object detection via vision and language knowledge distillation. arXiv preprint arXiv:2104.13921  (2021)

\bibitem{LVIS}
Gupta, A., Dollar, P., Girshick, R.: {LVIS}: A dataset for large vocabulary instance segmentation. In: CVPR (2019)

\bibitem{ResNet}
He, K., Zhang, X., Ren, S., Sun, J.: Deep residual learning for image recognition. In: CVPR (2016)

\bibitem{hu2018relation}
Hu, H., Gu, J., Zhang, Z., Dai, J., Wei, Y.: Relation networks for object detection. In: Proceedings of the IEEE conference on computer vision and pattern recognition. pp. 3588--3597 (2018)

\bibitem{AMSORT}
Kim, V., Jung, G., Lee, S.W.: Am-sort: Adaptable motion predictor with historical trajectory embedding for multi-object tracking. arXiv preprint arXiv:2401.13950  (2024)

\bibitem{TETer}
Li, S., Danelljan, M., Ding, H., Huang, T.E., Yu, F.: Tracking every thing in the wild. In: ECCV. Springer (2022)

\bibitem{OVTrack}
Li, S., Fischer, T., Ke, L., Ding, H., Danelljan, M., Yu, F.: Ovtrack: Open-vocabulary multiple object tracking. In: CVPR (2023)

\bibitem{masa}
Li, S., Ke, L., Danelljan, M., Piccinelli, L., Segu, M., Van~Gool, L., Yu, F.: Matching anything by segmenting anything. In: Proceedings of the IEEE/CVF Conference on Computer Vision and Pattern Recognition. pp. 18963--18973 (2024)

\bibitem{groundingdino}
Liu, S., Zeng, Z., Ren, T., Li, F., Zhang, H., Yang, J., Li, C., Yang, J., Su, H., Zhu, J., et~al.: Grounding dino: Marrying dino with grounded pre-training for open-set object detection. arXiv preprint arXiv:2303.05499  (2023)

\bibitem{TAO-OW}
Liu, Y., Zulfikar, I.E., Luiten, J., Dave, A., Ramanan, D., Leibe, B., O{\v{s}}ep, A., Leal-Taix{\'e}, L.: Opening up open world tracking. In: Proceedings of the IEEE/CVF Conference on Computer Vision and Pattern Recognition. pp. 19045--19055 (2022)

\bibitem{swin}
Liu, Z., Lin, Y., Cao, Y., Hu, H., Wei, Y., Zhang, Z., Lin, S., Guo, B.: Swin transformer: Hierarchical vision transformer using shifted windows. In: Proceedings of the IEEE/CVF international conference on computer vision. pp. 10012--10022 (2021)

\bibitem{cooler}
Liu, Z., Segu, M., Yu, F.: Cooler: Class-incremental learning for appearance-based multiple object tracking. In: DAGM German Conference on Pattern Recognition. pp. 443--458. Springer (2023)

\bibitem{TrackFormer}
Meinhardt, T., Kirillov, A., Leal-Taixe, L., Feichtenhofer, C.: Trackformer: Multi-object tracking with transformers. In: Proceedings of the IEEE/CVF conference on computer vision and pattern recognition. pp. 8844--8854 (2022)

\bibitem{MOT17}
Milan, A., Leal-Taix{\'e}, L., Reid, I., Roth, S., Schindler, K.: {MOT16}: A benchmark for multi-object tracking. arXiv preprint arXiv:1603.00831  (2016)

\bibitem{QDTrack}
Pang, J., Qiu, L., Li, X., Chen, H., Li, Q., Darrell, T., Yu, F.: Quasi-dense similarity learning for multiple object tracking. In: CVPR (2021)

\bibitem{CLIP}
Radford, A., Kim, J.W., Hallacy, C., Ramesh, A., Goh, G., Agarwal, S., Sastry, G., Askell, A., Mishkin, P., Clark, J., et~al.: Learning transferable visual models from natural language supervision. In: International conference on machine learning. pp. 8748--8763. PMLR (2021)

\bibitem{stablediffusion}
Rombach, R., Blattmann, A., Lorenz, D., Esser, P., Ommer, B.: High-resolution image synthesis with latent diffusion models. In: Proceedings of the IEEE/CVF conference on computer vision and pattern recognition. pp. 10684--10695 (2022)

\bibitem{supergule}
Sarlin, P.E., DeTone, D., Malisiewicz, T., Rabinovich, A.: Superglue: Learning feature matching with graph neural networks. In: Proceedings of the IEEE/CVF conference on computer vision and pattern recognition. pp. 4938--4947 (2020)

\bibitem{segu2024walker}
Segu, M., Piccinelli, L., Li, S., Van~Gool, L., Yu, F., Schiele, B.: Walker: Self-supervised multiple object tracking by walking on temporal appearance graphs. In: Computer Vision--ECCV 2024: 18th European Conference, Milan, Italy, September 29--October 4, 2024, Proceedings. Springer (2024)

\bibitem{segu2024samba}
Segu, M., Piccinelli, L., Li, S., Yang, Y.H., Schiele, B., Van~Gool, L.: Samba: Synchronized set-of-sequences modeling for end-to-end multiple object tracking. arXiv preprint  (2024)

\bibitem{darth}
Segu, M., Schiele, B., Yu, F.: Darth: holistic test-time adaptation for multiple object tracking. In: Proceedings of the IEEE/CVF International Conference on Computer Vision. pp. 9717--9727 (2023)

\bibitem{Waymo}
Sun, P., Kretzschmar, H., Dotiwalla, X., Chouard, A., Patnaik, V., Tsui, P., Guo, J., Zhou, Y., Chai, Y., Caine, B., et~al.: Scalability in perception for autonomous driving: Waymo open dataset. In: Proceedings of the IEEE/CVF conference on computer vision and pattern recognition. pp. 2446--2454 (2020)

\bibitem{TransTrack}
Sun, P., Cao, J., Jiang, Y., Zhang, R., Xie, E., Yuan, Z., Wang, C., Luo, P.: Transtrack: Multiple object tracking with transformer. arXiv preprint arXiv:2012.15460  (2020)

\bibitem{JDE}
Wang, Z., Zheng, L., Liu, Y., Wang, S.: Towards real-time multi-object tracking. The European Conference on Computer Vision (ECCV)  (2020)

\bibitem{DeepSORT}
Wojke, N., Bewley, A., Paulus, D.: Simple online and realtime tracking with a deep association metric. In: ICIP (2017)

\bibitem{GLEE}
Wu, J., Jiang, Y., Liu, Q., Yuan, Z., Bai, X., Bai, S.: General object foundation model for images and videos at scale. arXiv preprint arXiv:2312.09158  (2023)

\bibitem{IDOL}
Wu, J., Liu, Q., Jiang, Y., Bai, S., Yuille, A., Bai, X.: In defense of online models for video instance segmentation. ECCV  (2022)

\bibitem{UNINEXT}
Yan, B., Jiang, Y., Wu, J., Wang, D., Yuan, Z., Luo, P., Lu, H.: Universal instance perception as object discovery and retrieval. In: CVPR (2023)

\bibitem{ye2023cascade}
Ye, M., Ke, L., Li, S., Tai, Y.W., Tang, C.K., Danelljan, M., Yu, F.: Cascade-detr: delving into high-quality universal object detection. In: Proceedings of the IEEE/CVF International Conference on Computer Vision. pp. 6704--6714 (2023)

\bibitem{MOTR}
Zeng, F., Dong, B., Zhang, Y., Wang, T., Zhang, X., Wei, Y.: Motr: End-to-end multiple-object tracking with transformer. In: European Conference on Computer Vision. pp. 659--675. Springer (2022)

\bibitem{bytetrack}
Zhang, Y., Sun, P., Jiang, Y., Yu, D., Yuan, Z., Luo, P., Liu, W., Wang, X.: Bytetrack: Multi-object tracking by associating every detection box. In: ECCV (2022)

\bibitem{FairMOT}
Zhang, Y., Wang, C., Wang, X., Zeng, W., Liu, W.: {FairMOT}: On the fairness of detection and re-identification in multiple object tracking. IJCV  (2021)

\bibitem{nettrack}
Zheng, G., Lin, S., Zuo, H., Fu, C., Pan, J.: Nettrack: Tracking highly dynamic objects with a net. In: Proceedings of the IEEE/CVF Conference on Computer Vision and Pattern Recognition. pp. 19145--19155 (2024)

\bibitem{Detic}
Zhou, X., Girdhar, R., Joulin, A., Kr{\"a}henb{\"u}hl, P., Misra, I.: Detecting twenty-thousand classes using image-level supervision. In: ECCV (2022)

\end{thebibliography}
% \newpage
\section{Appendix}
In this supplementary material, we provide additional ablation studies and results of SLAck. We also elaborate on our experimental setup, method details, and training and inference hyper-parameters. 
All ablations are validated in the novel split of open-vocabulary MOT benchmark~\cite{OVTrack}.

\subsection{Compare Object Motions on Different Datasets}

\begin{figure}[!t]
    \centering
    \includegraphics[width=0.9\linewidth]{figs/SLAck_fastmotion_and_deformation_v2.pdf}
    \caption{Unlike conventionally pedestrian tracking such as MOT20~\cite{MOT20}, open-world objects show high dynamicity on shape and motion, which poses significant challenges for motion-based trackers. We demonstrate this quantitatively on ~Fig.~\ref{fig:kde_taovsmot}. }
    \label{fig:ov-tao-deformation}

    \begin{minipage}[t]{0.49\textwidth}
   \includegraphics[width=\linewidth]{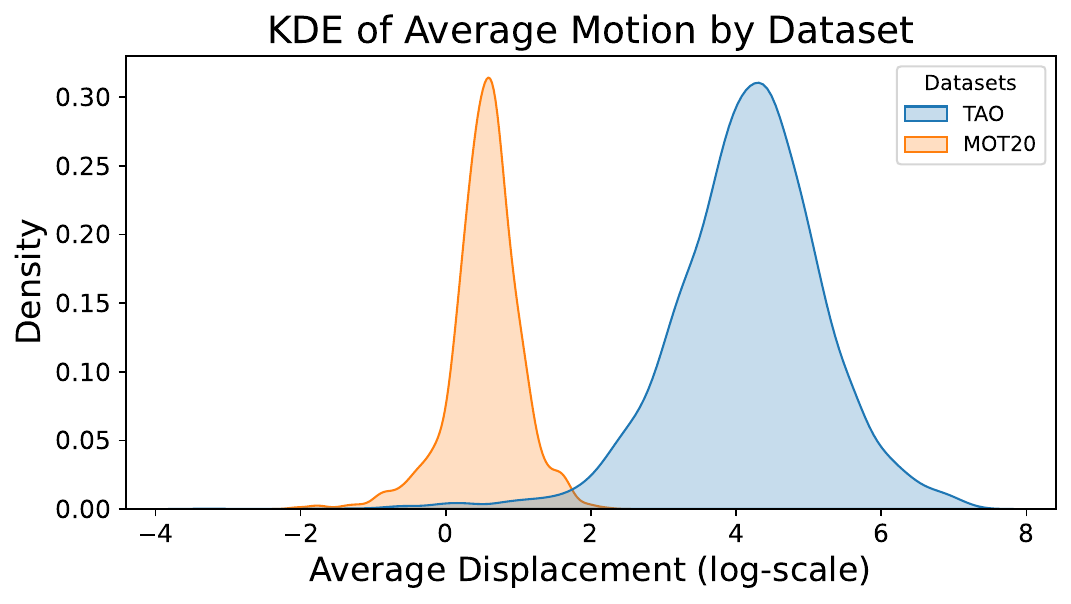}
        \label{fig:plot1}
    \end{minipage}\hfill
    \begin{minipage}[t]{0.49\textwidth}
        \includegraphics[width=\linewidth]{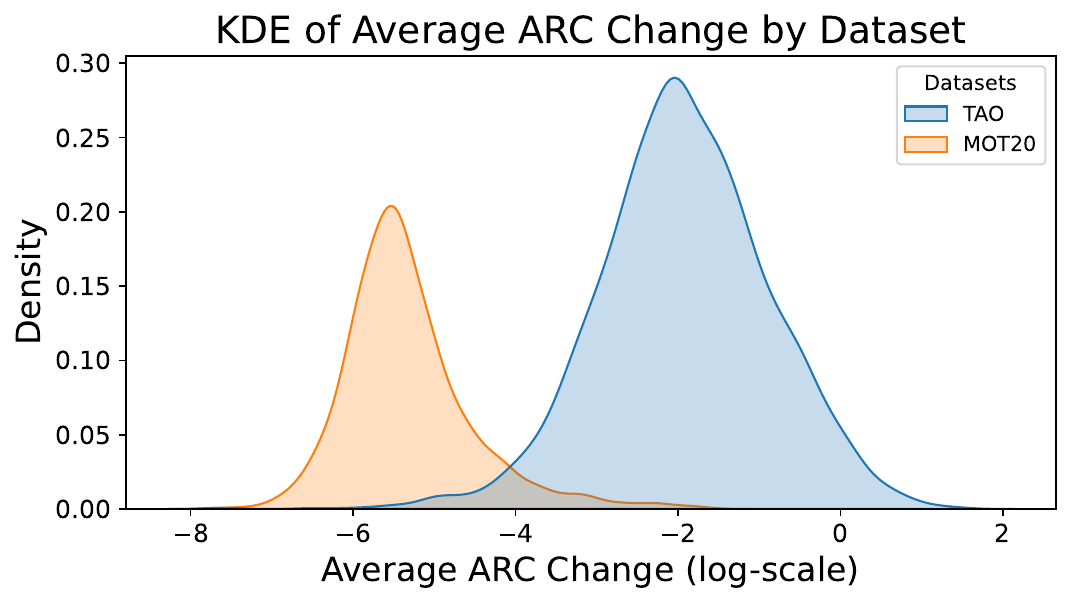}
    \end{minipage}

    \caption{Kernel density estimation (KDE) serves as a tool for comparing motion patterns between the open-vocabulary tracking dataset TAO and the conventional pedestrian tracking dataset MOT20. Given the substantial differences, we apply a logarithmic scale on the x-axis for both figures to facilitate a clearer comparison. \textbf{Left-KDE} illustrates the average motion disparities between objects in TAO and MOT20 datasets. For each instance, we compute the displacement between two consecutive frames. The open-vocabulary tracking dataset TAO showcases a broader variety of movements, as indicated by the width of the plot, and the average displacement is considerably larger than that observed in pedestrian tracking within MOT20. \textbf{Right KDE} focuses on the average Aspect Ratio Change (ARC) across the two datasets. This plot further confirms that, in open-vocabulary scenarios, objects undergo more pronounced deformation and occlusion, underscoring the complexities inherent in open-vocabulary tracking compared to traditional pedestrian tracking.
}
    \label{fig:kde_taovsmot}
\end{figure}

\label{sec:motion_datasets_difference}
In the main paper, we discuss the differences between pedestrian tracking and open-vocabulary tracking, which are also visually presented in Fig. \ref{fig:ov-tao-deformation}. Instances in pedestrian tracking datasets like MOT20\cite{MOT20} generally move in a linear path with minimal deformation. This simplicity allows many top-performing trackers to rely on the Kalman Filter. However, open-vocabulary tracking introduces significant challenges due to the complex, non-linear movements and substantial deformation of objects.

We also quantitatively showcase these differences in Fig. \ref{fig:kde_taovsmot} by plotting kernel density estimation (KDE) based on two aspects: object displacement between consecutive frames in a video and the average Aspect Ratio Changes (ARC) of instances throughout the video. We analyze and visualize the KDE distribution by examining all instance trajectories in the MOT20\cite{MOT20} and TAO~\cite{TAO} datasets.

To calculate object motion, we measure the displacement between two consecutive frames for each instance by computing the Euclidean distance between the centroids of their bounding boxes. For ARC, we determine the aspect ratio (width divided by height) of each bounding box per frame and then calculate the changes in aspect ratio between consecutive frames for each instance.

Fig. \ref{fig:kde_taovsmot} further supports our qualitative observation from Fig. \ref{fig:ov-tao-deformation} that, compared to the traditional pedestrian tracking dataset MOT20, objects in open-vocabulary settings move with higher dynamism in both motion and shape. This is indicated by the broader KDE curve and the significantly larger average values for instance displacement and ARC changes.

\subsection{Semantics and Motion Patterns}
\begin{figure}[ht]
    \centering
        \includegraphics[width=\linewidth]{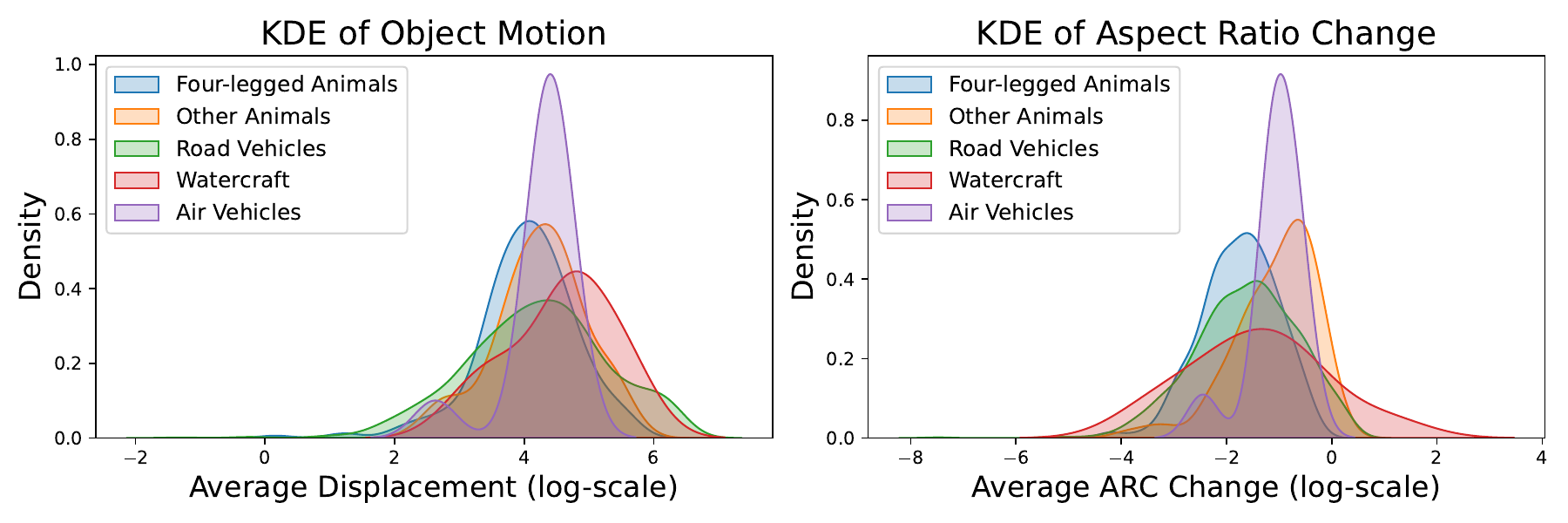}
    \caption{Kernel density estimation (KDE) for comparing motion patterns in different classes for on the TAO dataset.}
    \label{fig:parent_class_kde}
\end{figure}

\begin{figure}[t]
    \centering
    \includegraphics[width=0.8\linewidth]{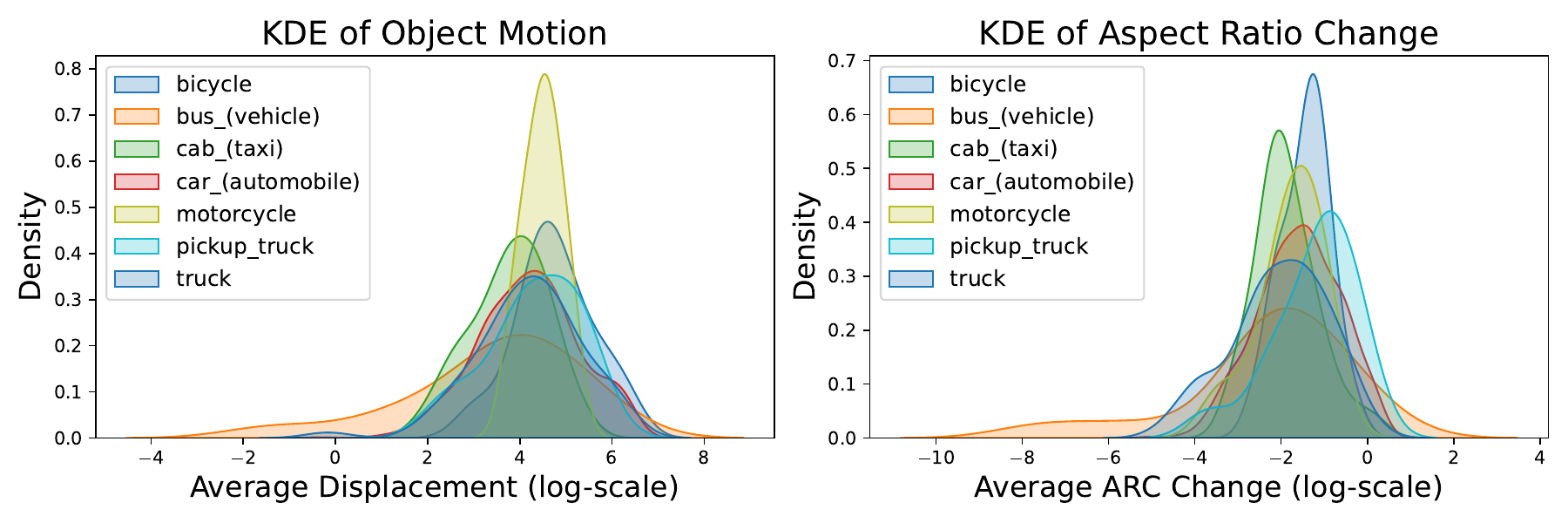}
    \caption{The KDE of motion displacement and ARC for different classes in road vehicles.}
    \label{fig:road_vehicle_kde}

    \includegraphics[width=0.8\linewidth]{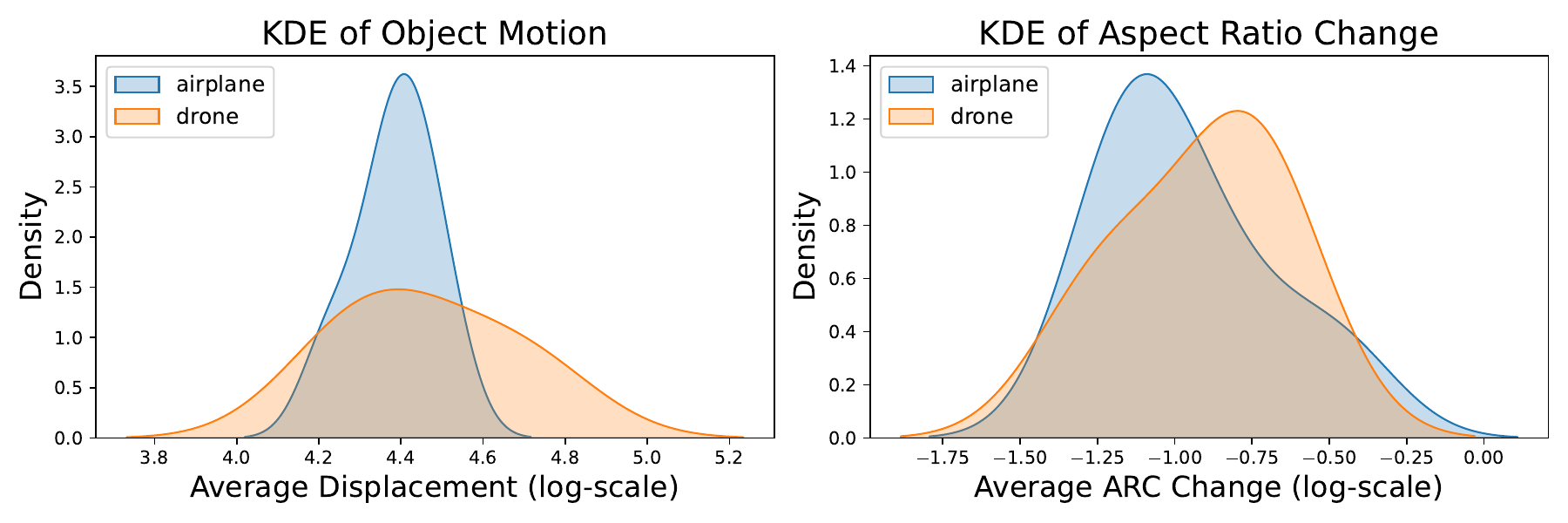}
    \caption{The KDE of motion displacement and ARC for different classes in air vehicles.}
    \label{fig:air-vehicle-kde}

    \includegraphics[width=0.8\linewidth]{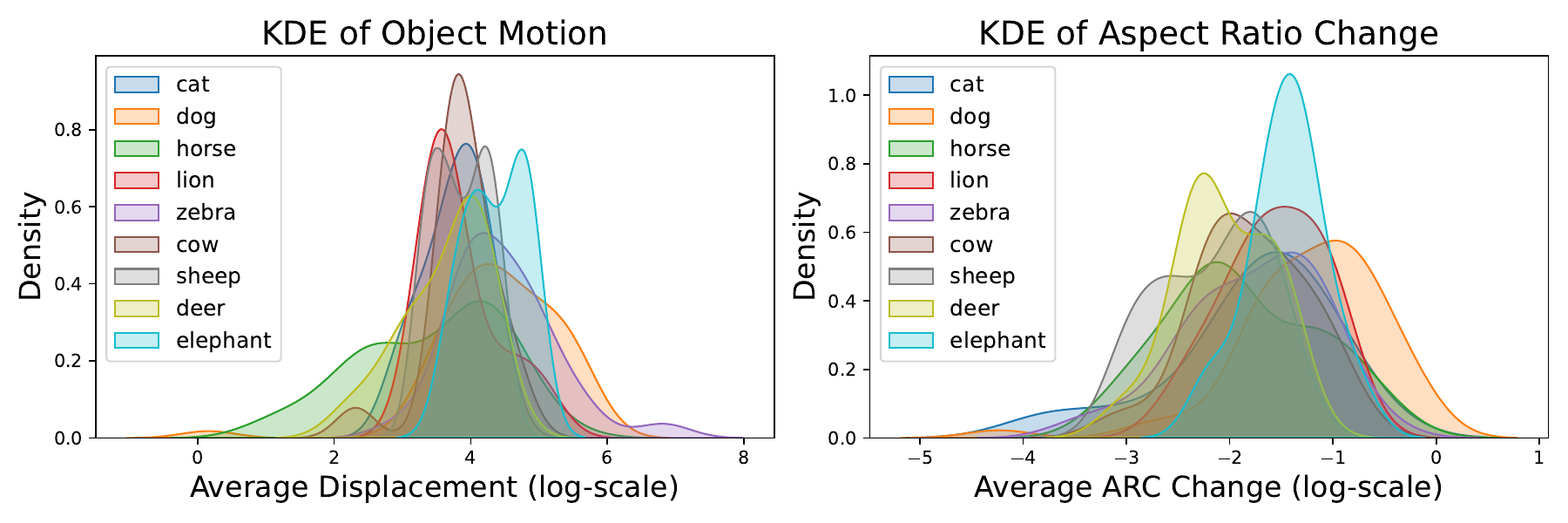}
    \caption{The KDE of motion displacement and ARC for different classes in four-leg animals.}
    \label{fig:four-leg-animal-kde}

    \includegraphics[width=0.8\linewidth]{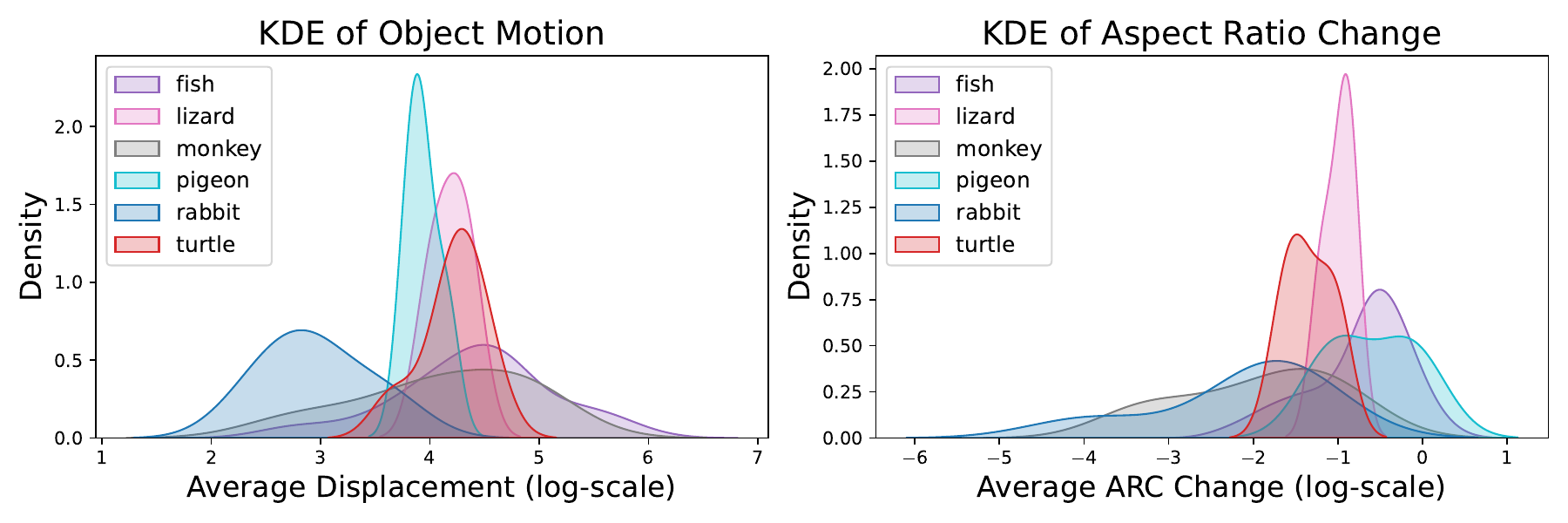}
    \caption{The KDE of motion displacement and ARC for different classes in other animals.}
    \label{fig:other-animal-kde}
\end{figure}

In the main paper, we explain the effectiveness of SLAck over previous state-of-the-art open-vocabulary trackers that rely solely on appearance information. Our approach, which integrates semantic and location information, is based on the understanding that objects from different classes exhibit unique motion patterns. We now provide a quantitative demonstration using two descriptors for object motion defined earlier: average displacement and average aspect ratio changes. While this simplification doesn't capture all motion nuances, it offers a straightforward method for analyzing and visualizing motion patterns across different classes.

We calculate the average displacement and aspect ratio change (ARC) for each instance, then aggregate these metrics by class to uncover general motion characteristics and offer insights into the frequency and severity of deformation or occlusion for each class.

\noindent{}\textbf{Semantic and Motion Relations for Parent Classes}
We categorize TAO classes into three main groups: Transport, Animal, and Static objects, focusing only on Transport and Animals—objects capable of autonomous movement. Static objects, which include items that must be carried or transported by others (e.g., personal items, sports equipment, household items, and work tools), are not analyzed due to their inherent lack of independent motion.

\begin{itemize}
    \item \textbf{Transport} encompasses all means of transportation, organized by similarity in motion patterns. This includes road vehicles with similar motion patterns, air vehicles like airplanes and drones, and watercraft such as boats.
    \item \textbf{Animals} are grouped based on their locomotion methods or habitats, emphasizing similar motion patterns (e.g., four-legged land animals, birds, and fish).
\end{itemize}

Results for parent classes are shown in Fig.~\ref{fig:parent_class_kde}, with one figure illustrating the motion displacement density estimation and the other showing the average ARC for different categories. These figures should be considered together to fully understand the motion patterns and specific characteristics of each category. As illustrated, the distribution of motion patterns varies significantly across parent classes, confirming our assumption that objects from different classes move differently.

\noindent{}\textbf{Semantic and Motion Relations for Child Classes}
Delving deeper, we examine the child categories within each parent group to analyze more specific motion patterns. Through KDE analysis, we present detailed findings for subclasses, such as road vehicles (Fig. \ref{fig:road_vehicle_kde}), air vehicles (Fig. \ref{fig:air-vehicle-kde}), four-legged animals (Fig. \ref{fig:four-leg-animal-kde}), and other animals (Fig. \ref{fig:other-animal-kde}). Our observations reveal that even within semantically similar classes, distinct motion pattern variations exist. SLAck leverages these differences by learning the relationships between semantic categories and their associated motion patterns directly from the data, enhancing its tracking accuracy.

\subsection{More Ablations}
We provide more ablation studies in this section.
\begin{table}[!t]
  \centering
  \resizebox{0.9\textwidth}{!}{
    \scriptsize
    \begin{minipage}{0.33\linewidth}
      \centering
      \caption{Sampling interval.}
      \begin{tabular}{@{}l|cc@{}}
        \toprule
        & Novel & Base \\
        \cmidrule(r){2-3}
        Time & AssocA & AssocA \\
        \midrule
        1s  & 33.4 & 36.7 \\
        3s  & \textbf{37.8} & \textbf{37.6} \\
        5s  & 35.7 & 36.5 \\
        10s & 35.7 & 35.5 \\
        20s & 34.2 & 33.8 \\
        \bottomrule
      \end{tabular}
      \label{tab:time_inteval}
    \end{minipage}
    \hfill
    \begin{minipage}{0.33\linewidth}
      \centering
      \caption{Concat vs Add.}
      \begin{tabular}{@{}l|cc@{}}
        \toprule
        & Novel & Base \\
        \cmidrule(r){2-3}
        Model & AssocA & AssocA \\
        \midrule
        Concat & 35.8 & \textbf{37.7} \\
        Add & \textbf{37.8} & {37.6} \\
        \bottomrule
      \end{tabular}
      \label{tab:concat_vs_add}
    \end{minipage}
    \hfill
    \begin{minipage}{0.33\linewidth}
      \centering
      \caption{img vs text.}
      \begin{tabular}{@{}l|cc@{}}
        \toprule
        & Novel & Base \\
        \cmidrule(r){2-3}
        Model & AssocA & AssocA \\
        \midrule
        img  & 35.9 & 35.8 \\
        img + text  & 37.8 & 36.5 \\
        text  & \textbf{37.8} & \textbf{37.6} \\
        \bottomrule
      \end{tabular}
      \label{tab:img_vs_text}
    \end{minipage}
  }
\end{table}

\begin{table}[t!]
  \centering
  \caption{Comparison with SOT method on TAO novel split.}
  \resizebox{0.7\textwidth}{!}{
    \scriptsize
    \begin{tabular}{@{}l|ccccr@{}}
      \toprule
      & TETA & LocA & AssocA & ClsA & Inference Time \\
      \midrule
      MixFormer & 20.6 & 31.2 & 28.4 & \textbf{2.4} & 36h \\
      SlAck (Ours) & \textbf{31.1} & \textbf{54.3} & \textbf{37.8} & 1.3 & \textbf{1.2h} \\
      \bottomrule
    \end{tabular}
  }
  \label{tab:sot_res}
\end{table}

\subsection{Temporal Sampling Intervals}
We further perform ablation for temporal sampling interval, as shown in Table~\ref{tab:time_inteval}. 
Longer intervals risk objects disappearing, while shorter intervals fail to capture valuable cases such as occlusion. 

\subsection{Add vs. Concat for Feature Fusion}

We compare addition against concatenation to fuse different cues in Table~\ref{tab:concat_vs_add}. Results show that addition yields better performance on novel classes compared to concatenation. Addition maintains fixed-size dimensionality, whereas concatenation increases dimensionality, leading to higher memory consumption and slower processing. 

\subsection{Distilled CLIP Features}

 We integrate the distilled CLIP text head features as semantic cues in the main paper. We here evaluated the integration of both distilled CLIP image and text head features in ~\cite{VILD}. Results in Table~\ref{tab:img_vs_text} show that using only the distilled CLIP text head features as semantic cues yields the best performance.

\subsection{Dynamic Thresholding}
\label{sec:dy_thr}
\begin{table}[t]
    \caption{Inference with or without dynamic thresholding (DT). }%
    \centering%
    \resizebox{0.5\linewidth}{!}{\begin{tabular}{l|cccc}
\toprule
Dynamic Thresholding & TETA & LocA & AssocA & ClsA \\
\midrule
OVTrack \textit{w/o} DT & 27.8 & 48.3 & 33.6 & 1.5 \\
Slack \textit{w/o} DT & 29.5 & 50.5 & 36.8 & 1.0 \\ \midrule
OVTrack \textit{w/} DT & 28.8 & 51.2 & 33.8 & 1.5 \\
Slack \textit{w/} DT & 31.1 & 54.3 & 37.8 & 1.3 \\
\bottomrule
\end{tabular}
}%
    \label{tab:dy_thre}%
\end{table}
We build SLAck with the same open-vocabulary detector used by OVTrack as a base. Our findings indicate that employing a fixed score threshold for open-vocabulary detection inference is suboptimal. This issue stems from the inherent design of FasterRCNN-based open-vocabulary detectors, which substitute the original classification head with a CLIP-distilled classification head. This setup calculates feature similarities among class text embeddings produced by the CLIP text encoder, followed by the softmax function to determine classification confidence for each class. Due to the introduction of novel classes during inference, the number of categories alters from the training phase, affecting the value distribution post-softmax. While OVTrack initially sets the score threshold during testing at 0.0001, our approach of utilizing a dynamic threshold, adjusted based on the number of target classes, yields improved detection outcomes. The test score threshold is determined by the following equation:

\begin{equation}
    \text{score\_thr} = \left(\frac{1}{\text{number of classes}}\right) \times 1.001
\end{equation}

We report these findings in Table~\ref{tab:dy_thre}, demonstrating that this modification leads to a +2.9 LocA enhancement for OVTrack and a +3.1 LocA increase for SLAck. In Table 4 of the main paper, we referenced the OVTrack performance on the standard benchmark without incorporating our insights. Herein, we also apply dynamic thresholding (DT) to OVTrack and juxtapose it with SLAck under both with and without DT scenarios. Table~\ref{tab:dy_thre} illustrates that SLAck surpasses OVTrack in both conditions, achieving at least a 3.2 improvement in association (AssocA).

\subsection{Temporal Encoding}
\begin{table}[t]
    \caption{With or without temporal encoding.}%
    \centering%
    \resizebox{0.7\linewidth}{!}{\begin{tabular}{l|cccc}
\toprule
Method & TETA & LocA & AssocA & ClsA \\
\midrule
SLAck \textit{w/o} temporal encoding & 30.6 & \textbf{54.4} & 36.8 & 0.7 \\
SLAck \textit{w/} temporal encoding   & \textbf{31.1} & {54.3} & \textbf{37.8} & \textbf{1.3} \\
\bottomrule
\end{tabular}
}%
    \label{tab:temporal_encoding}%
\end{table}

SLAck innovatively combines three crucial cues for the association: semantics, location, and appearances. Due to space constraints, the detailed explanation of temporal encoding was previously omitted. In this section, we elucidate temporal encoding and assess its efficacy. Temporal encoding aims to capture the differences between pairs of frames to be matched. For each frame, a unique encoding is generated for the entire frame. This is accomplished by downsampling the smallest feature maps in the Feature Pyramid Network (FPN) of the detector into a two-dimensional vector. Prior to matching objects across two frames, we compute the difference between the encodings of these frames, which constitutes the temporal encoding. This encoding reflects the changes between the two frames; if no movement occurs and the scene remains static, the difference approaches zero.

We replicate the temporal encoding for the same number of objects in the current frame and combine it with each object's fused embedding:

\begin{equation}
    E^i_{\text{fused\_update}} = E^i_{\text{fuse}} + E^i_{\text{temporal}},
\end{equation}

where \(E^i_{\text{fused}}\) denotes the fused embedding for the \(i\)-th object in the frame, and \(E^i_{\text{temporal}}\) represents the temporal encoding. Table~\ref{tab:temporal_encoding} demonstrates the temporal encoding's effectiveness, contributing to a +1 increase in AssocA for SLAck.

\subsection{Compare with Single Object Tracking (SOT)}
The significance of OV-MOT is its ability to support both open-vocabulary and \textbf{multiple objects}, whereas SOT focuses solely on tracking a single object.
In OV-MOT, objects frequently appear and disappear, and new objects continually enter the scene, necessitating automatic handling of these scenarios, which SOT trackers cannot achieve.
Additionally, OV-MOT often involves multiple objects with similar appearances appearing together, necessitating the use of motion or location cues.
Moreover, OV-MOT tracks all objects at once without repeated runs as needed in SOT.
Also, SOT fails to explicitly consider other objects in the scene, such as their semantics, relative location, and appearance that play a critical role in associating multiple objects. 
We test a strong SOT tracker MixFormer (GOT-10k) for the OV-MOT task on TAO in Table~\ref{tab:sot_res}. Our approach significantly outperforms MixFormer. We use the same detector as ours to provide initial object boxes for MixFormer in the first frame of each video and run it repeatedly to track every object.

\subsection{Model Details}

\noindent\textbf{Semantic Head}
The semantic head is a five-layer MLP with the GroupNorm and ReLU after each but the last layer.

\noindent\textbf{Location Head}
The semantic head is a five-layer MLP with the GroupNorm and ReLU after each but the last layer.

\noindent\textbf{Appearance Head}
The appearance head is a four-layer convolution with one additional MLP. After each conv layer follows a GroupNorm and ReLU.

\noindent\textbf{Spatial and Temporal Object Graph (STOG)}
STOG is designed to process and refine fused features through a series of attentional propagation layers. It employs a descriptor dimension of 256. The network's architecture consists of alternating layers of self-attention and cross-attention mechanisms, comprising four layers in total. Each layer leverages multi-head attention with 4 heads. This is followed by feature fusion, where the attention-derived features are concatenated with the original input features and subsequently refined through an MLP with layers configured as \texttt{[512, 512, 256]}. This MLP serves to project the enhanced feature vectors back to the original feature dimension, ensuring consistency across layers.

\subsection{Training Detail}
For SLAck-OV, we train our model using the same FasterRCNN-based open-vocabulary detector as OVTrack~\cite{OVTrack}. We conduct an inference of the pre-trained detector on pairs of sampled images using a dynamic score threshold described in Sec.~\ref{sec:dy_thr}. The maximum number of detection predictions per image is set to 50. Instead of the default intra-class non-maximum suppression (NMS), we apply class-agnostic NMS with an IoU threshold of 0.5. Following inference, we match the detection boxes with the sparse TAO ground truth using an IoU threshold of 0.7 for match determination. The matched detection boxes are then assigned instance IDs for association learning. It is possible for one ground truth box to match multiple detection boxes. During our differentiable optimal transport optimization using the Sinkhorn algorithm, we adjust the marginal distributions based on the sum of matches with ground truth. The number of Sinkhorn iterations is set to 100 during training. 

Training images are randomly resized, keeping the aspect ratio, with the shorter side between 640 and 800 pixels. Pairs of adjacent frames, within a maximum interval of 3 seconds, are selected for training. Models are trained for 12 epochs using a batch size of 16, applying an SGD optimizer with an initial learning rate of 0.008 and weight decay of 0.0001.
For SLAck-T and SLAck-L, all the hyper-parameters described above are the same as SLAck-OV except that we use a fixed score threshold of 0.0001 for the detection inference in detection-aware training.

\begin{figure*}[t]
    \centering
    \small
    \setlength\tabcolsep{0.5mm}
    \resizebox{1.0\linewidth}{!}{
    \begin{tabular}{cccc}
        \toprule
        $t$ & $t +1$ & $t + 2$& $t + 3$\\ \midrule

        \includegraphics[width=0.2\linewidth]{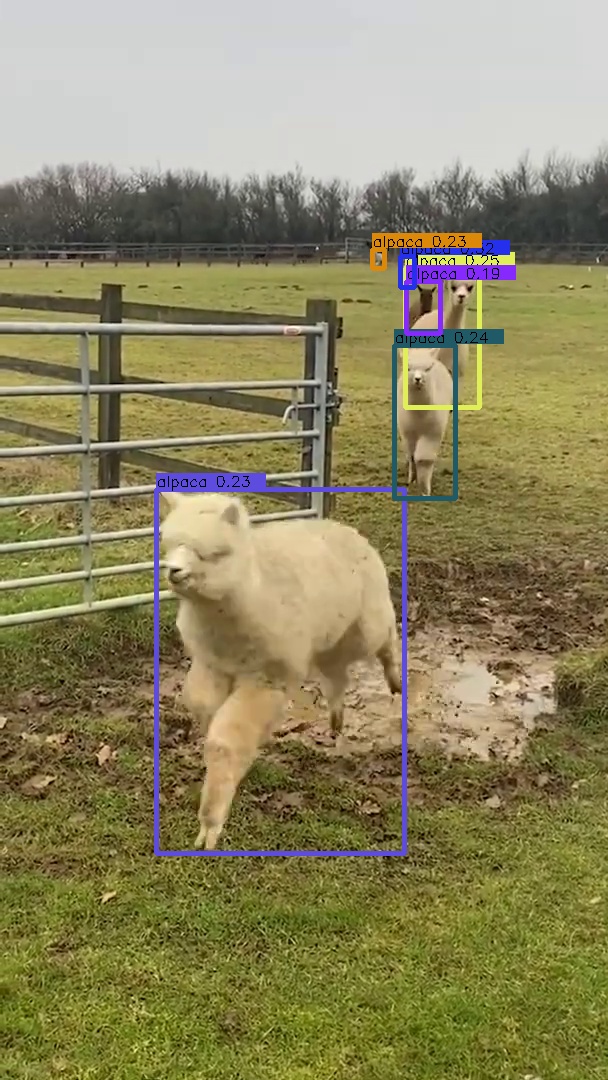} &
        \includegraphics[width=0.2\linewidth]{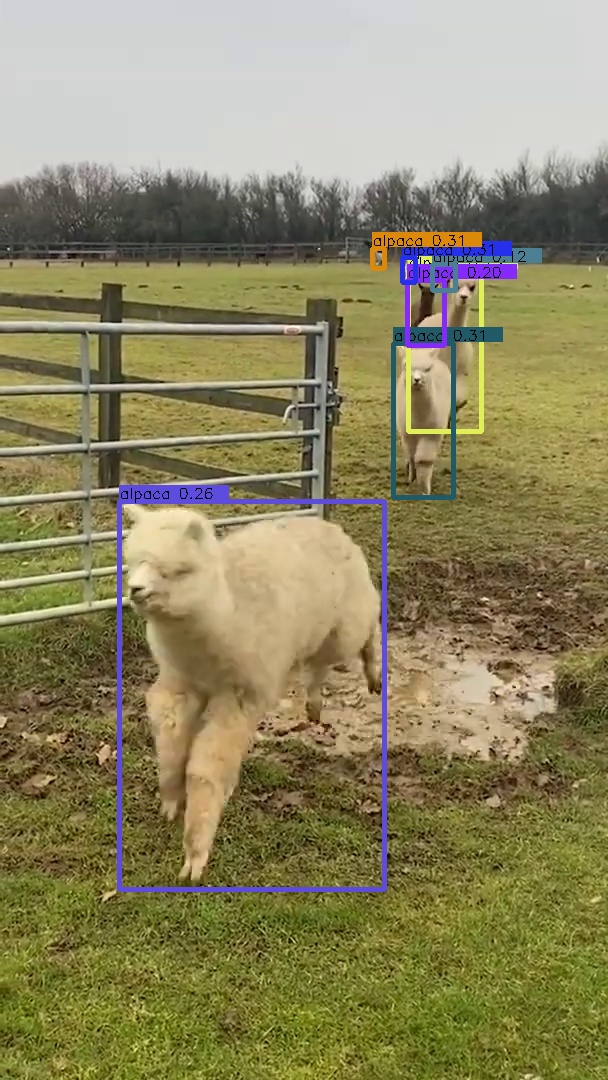} &
        \includegraphics[width=0.2\linewidth]{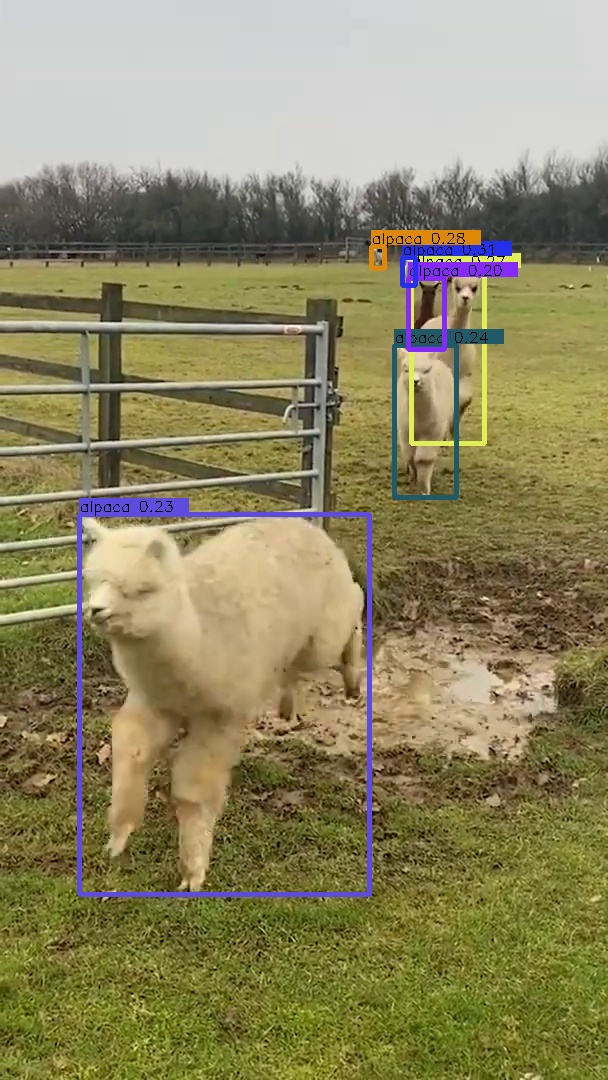} &
        \includegraphics[width=0.2\linewidth]{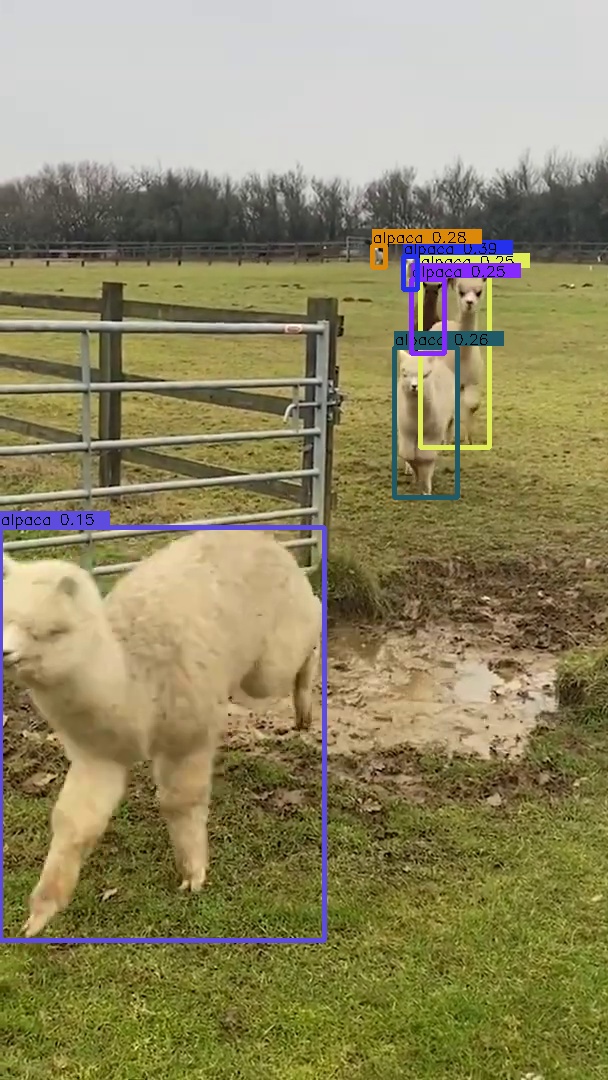} \\
        
        \includegraphics[width=0.2\linewidth]{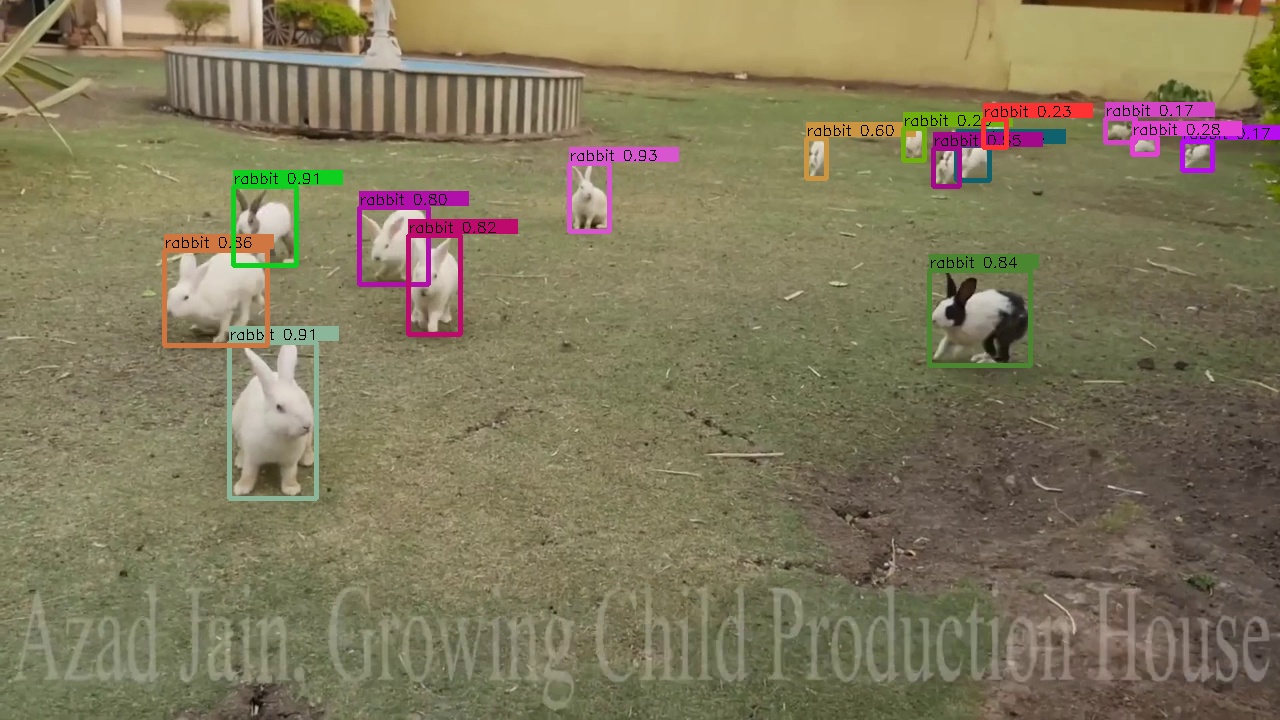} &
        \includegraphics[width=0.2\linewidth]{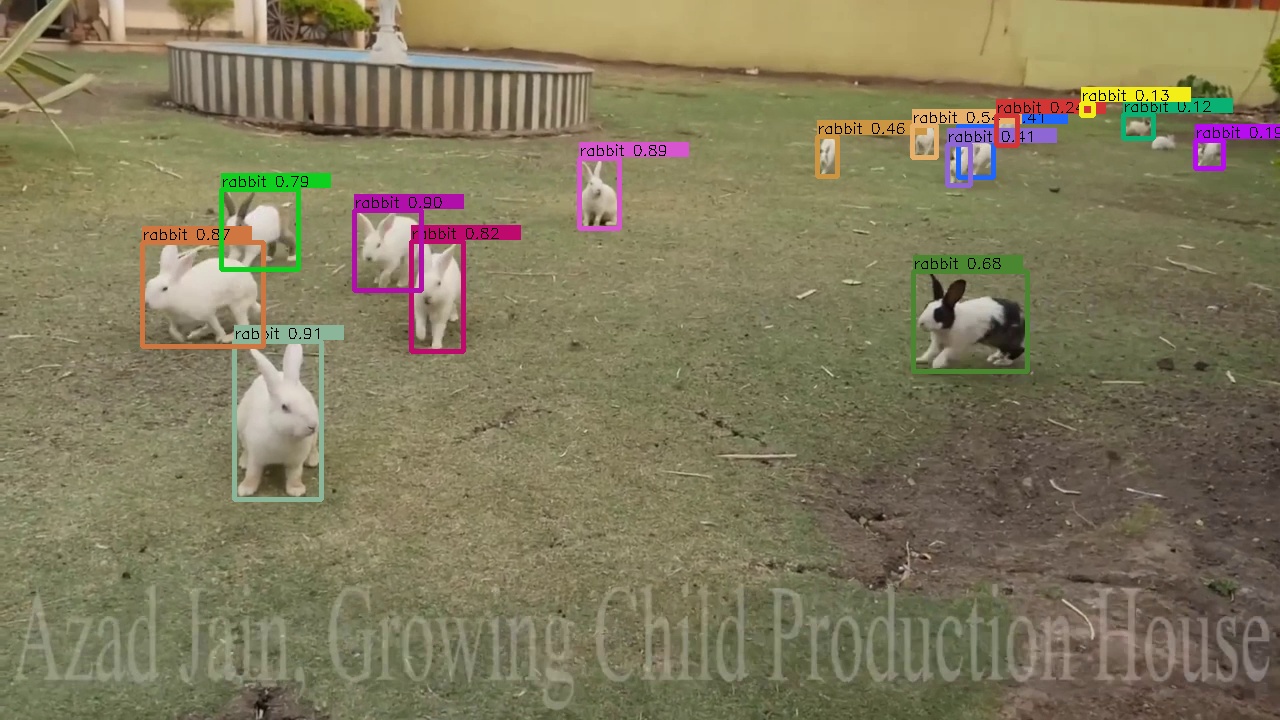} &
        \includegraphics[width=0.2\linewidth]{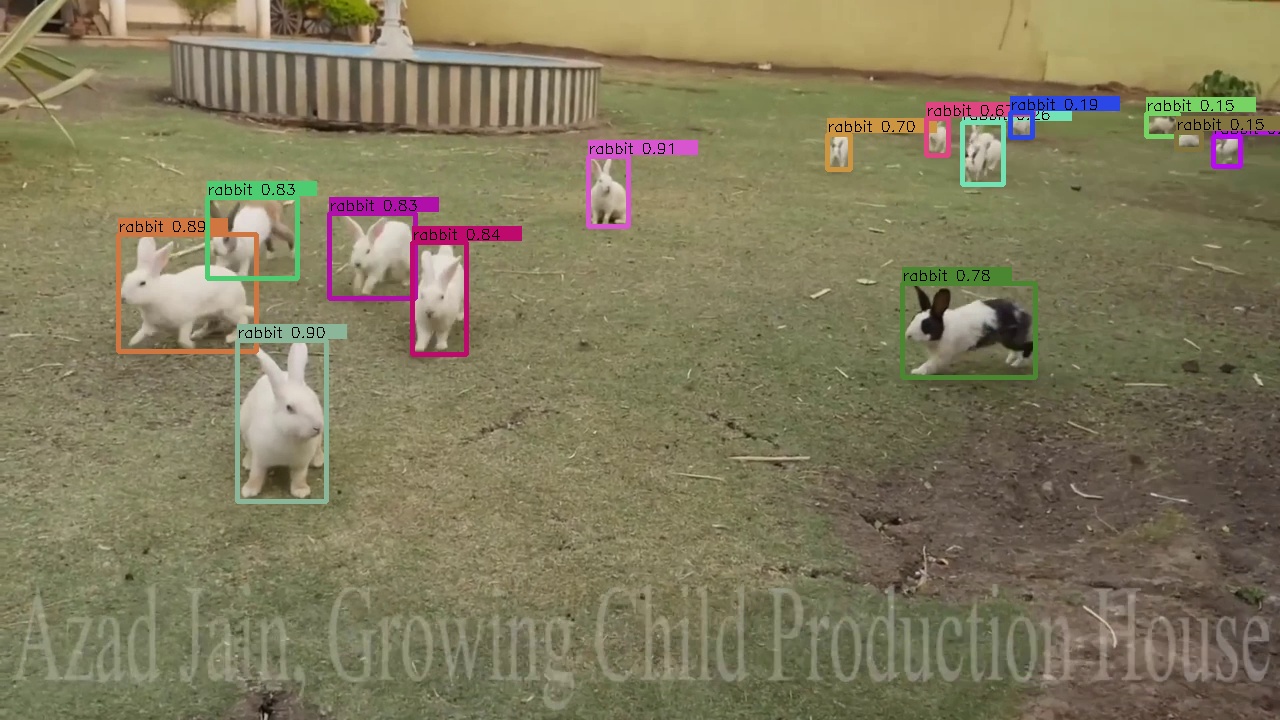} &
        \includegraphics[width=0.2\linewidth]{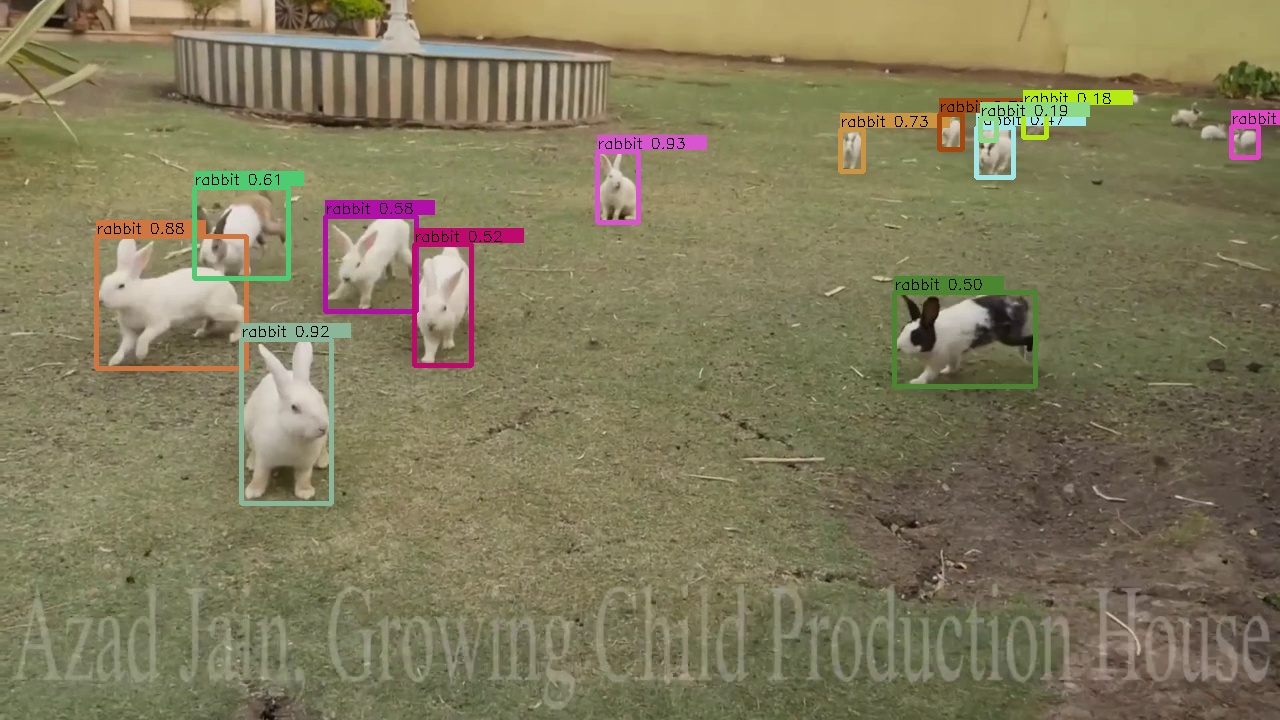} \\
        
        \includegraphics[width=0.2\linewidth]{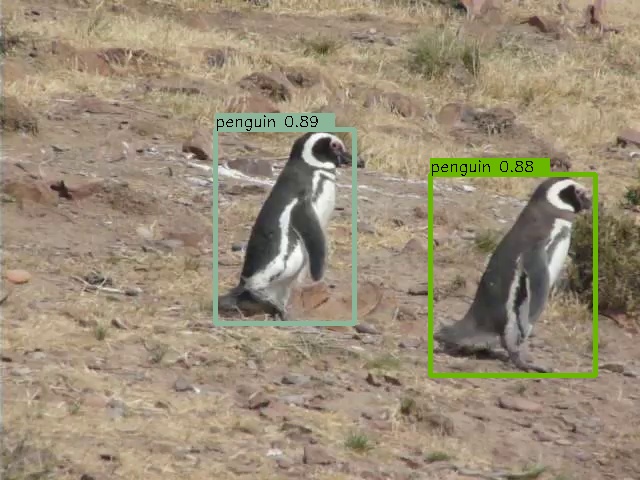} &
        \includegraphics[width=0.2\linewidth]{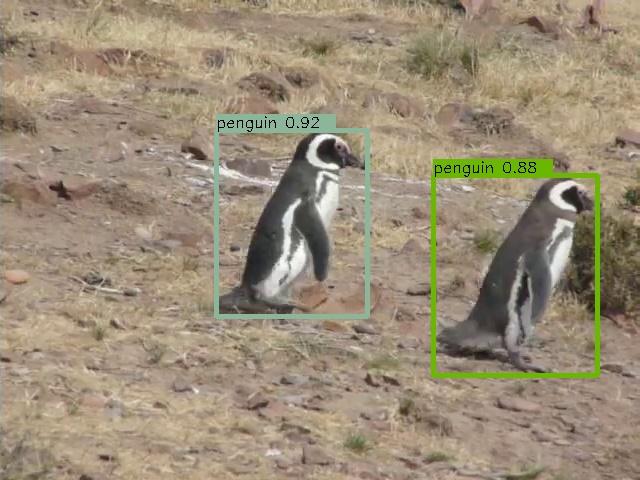} &
        \includegraphics[width=0.2\linewidth]{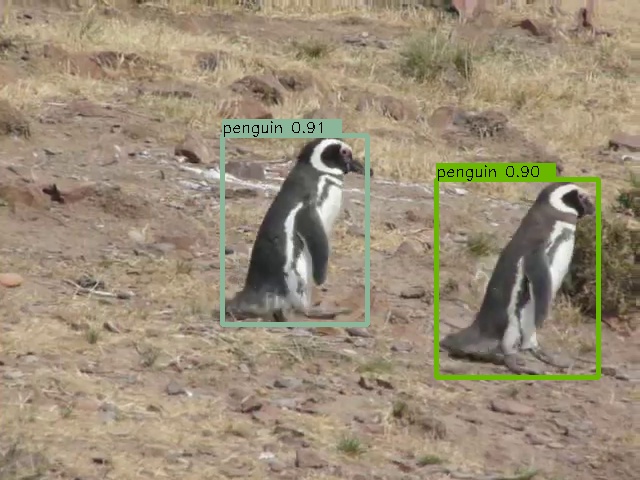} &
        \includegraphics[width=0.2\linewidth]{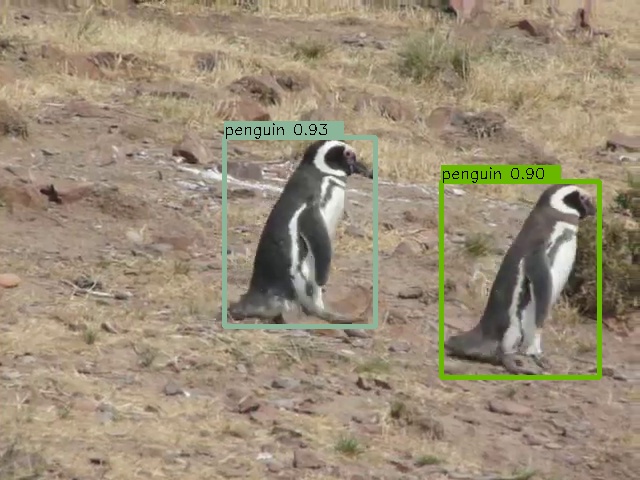} \\
   
        \includegraphics[width=0.2\linewidth]{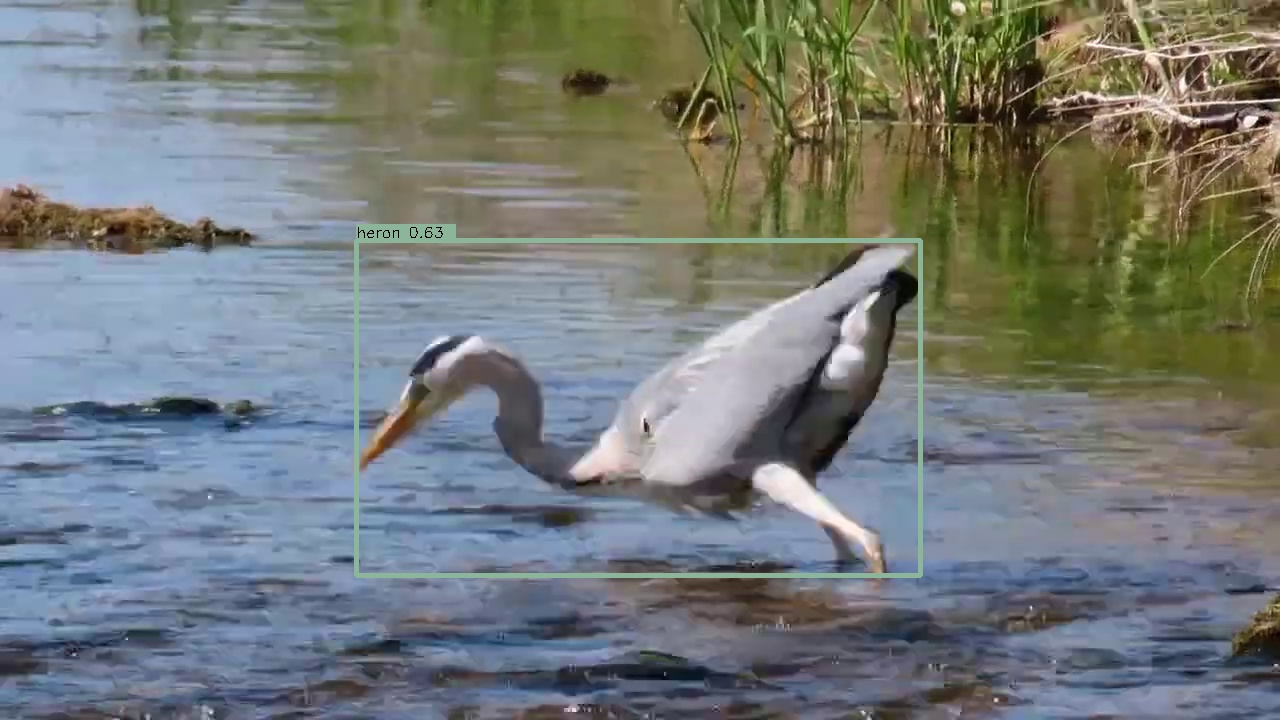} &
        \includegraphics[width=0.2\linewidth]{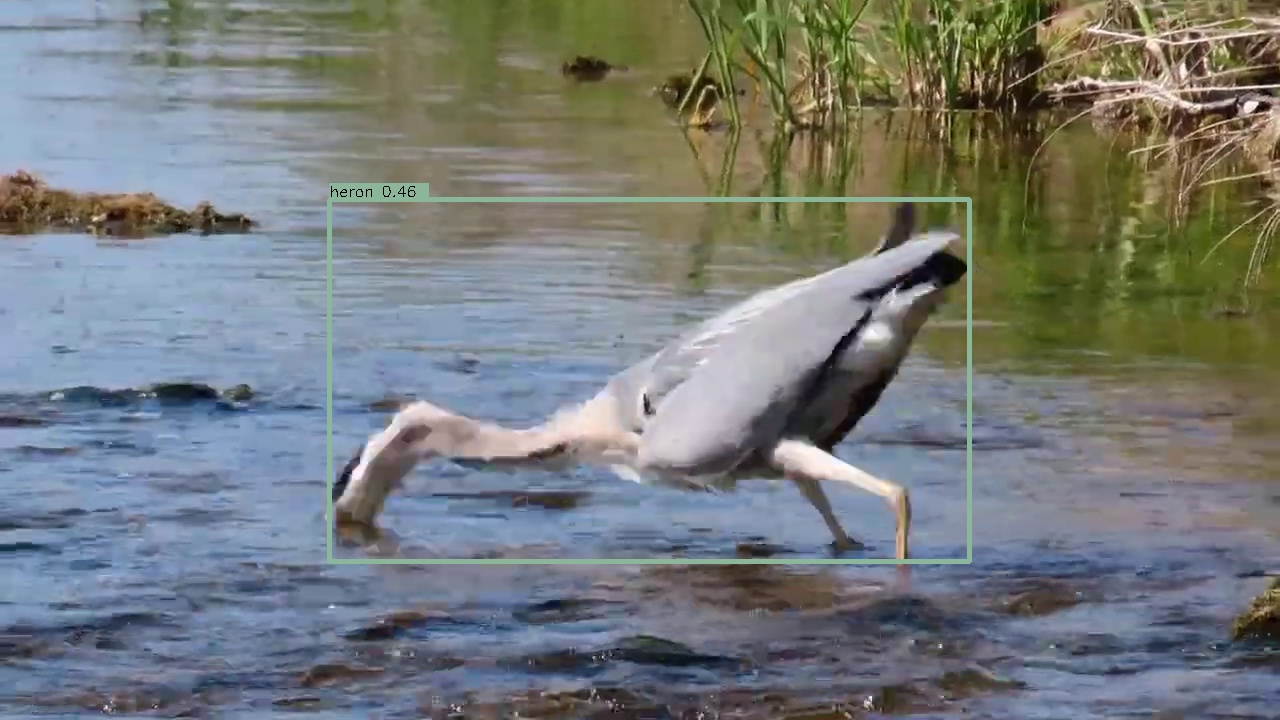} &
        \includegraphics[width=0.2\linewidth]{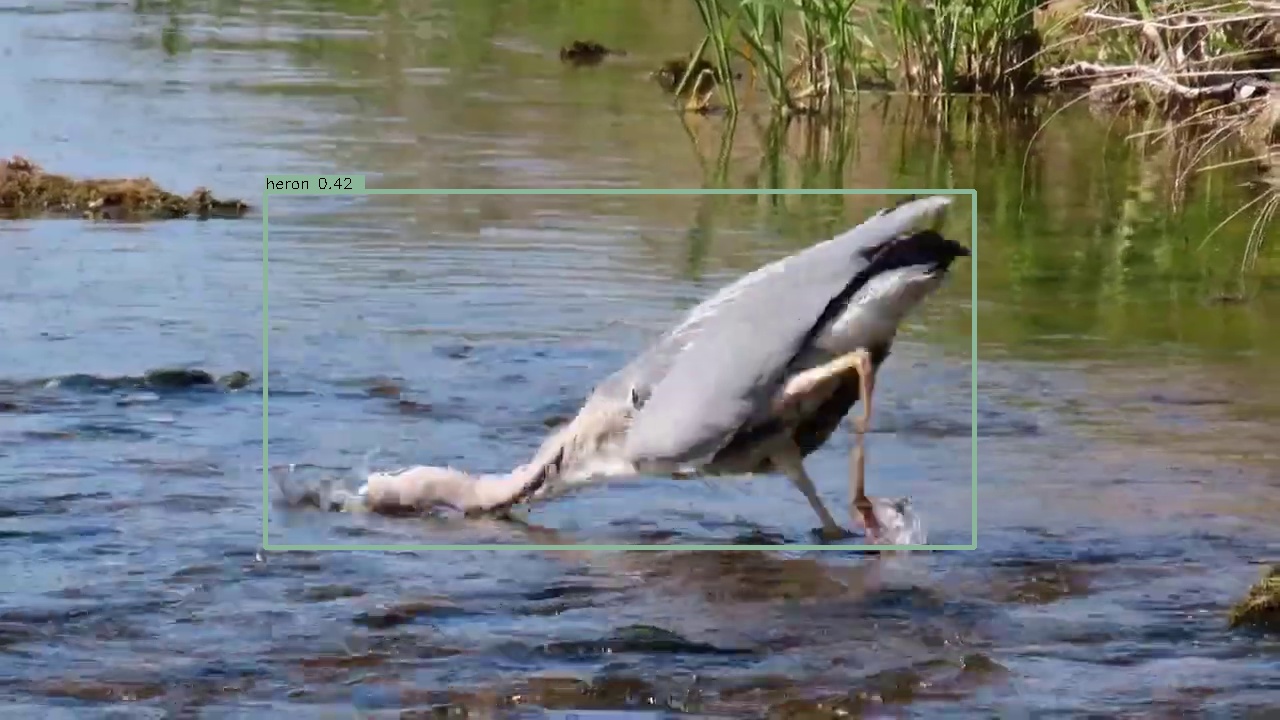} &
        \includegraphics[width=0.2\linewidth]{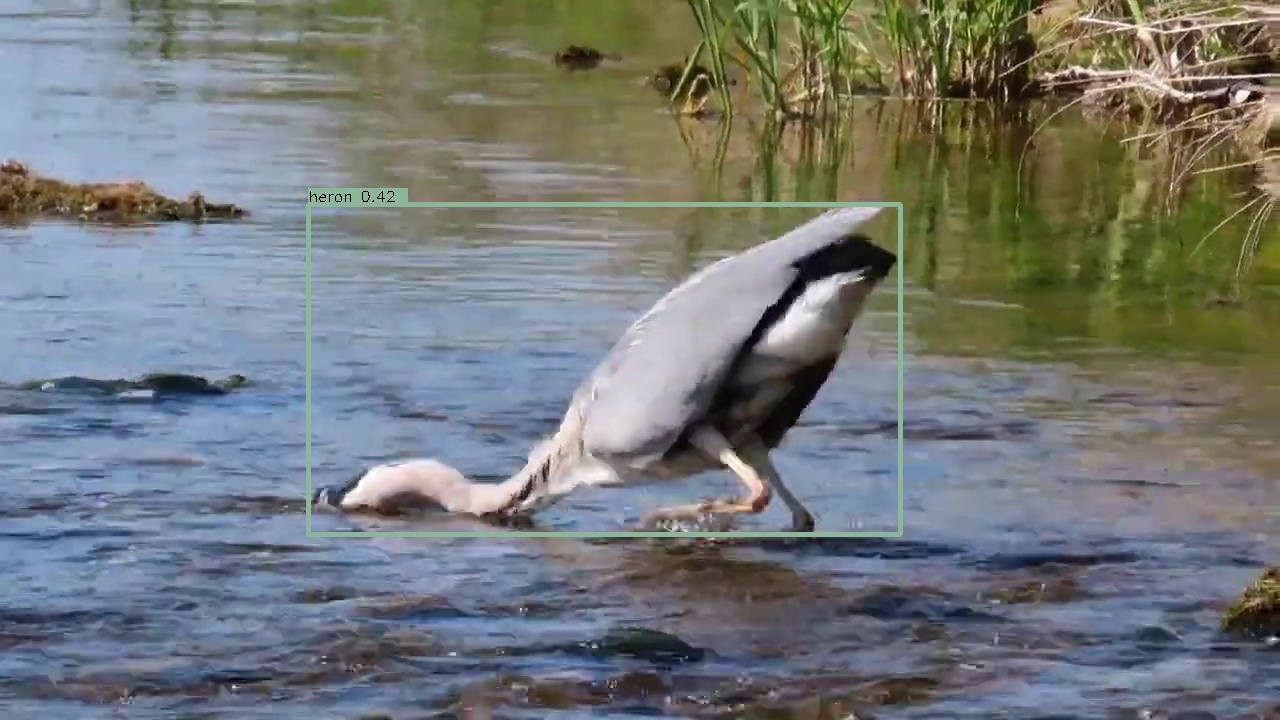} \\
        
        \includegraphics[width=0.2\linewidth]{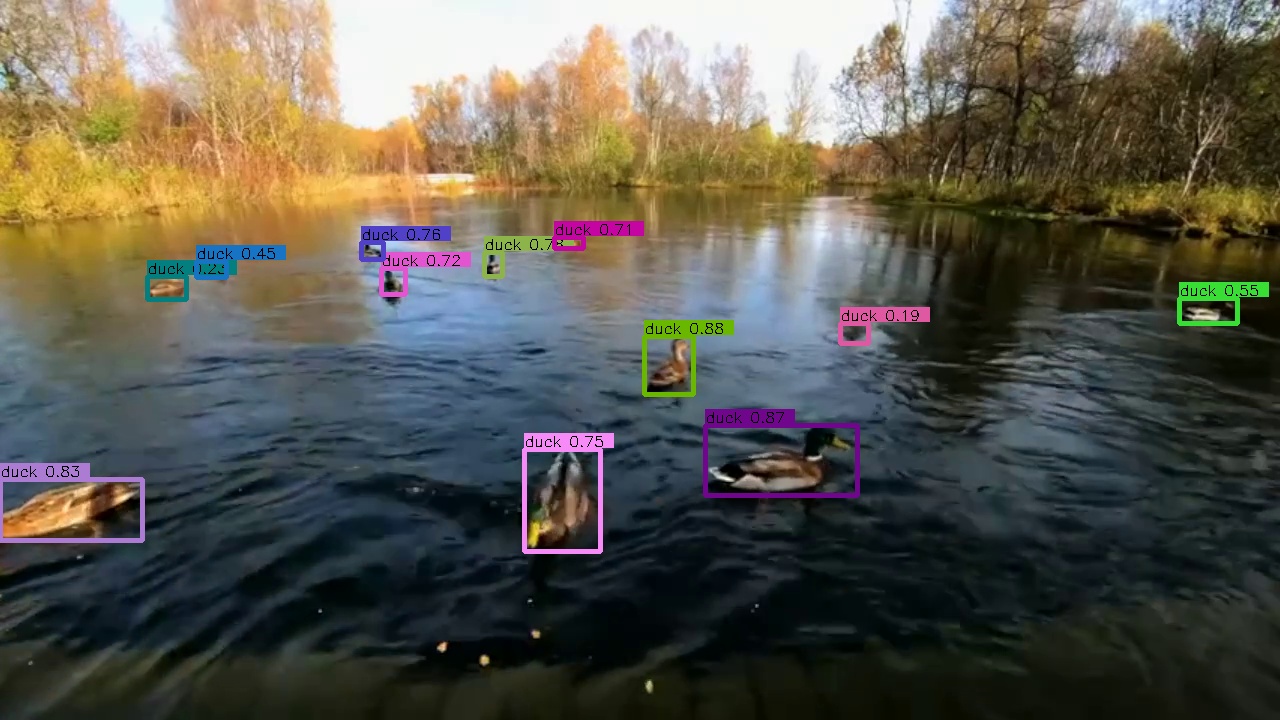} &
        \includegraphics[width=0.2\linewidth]{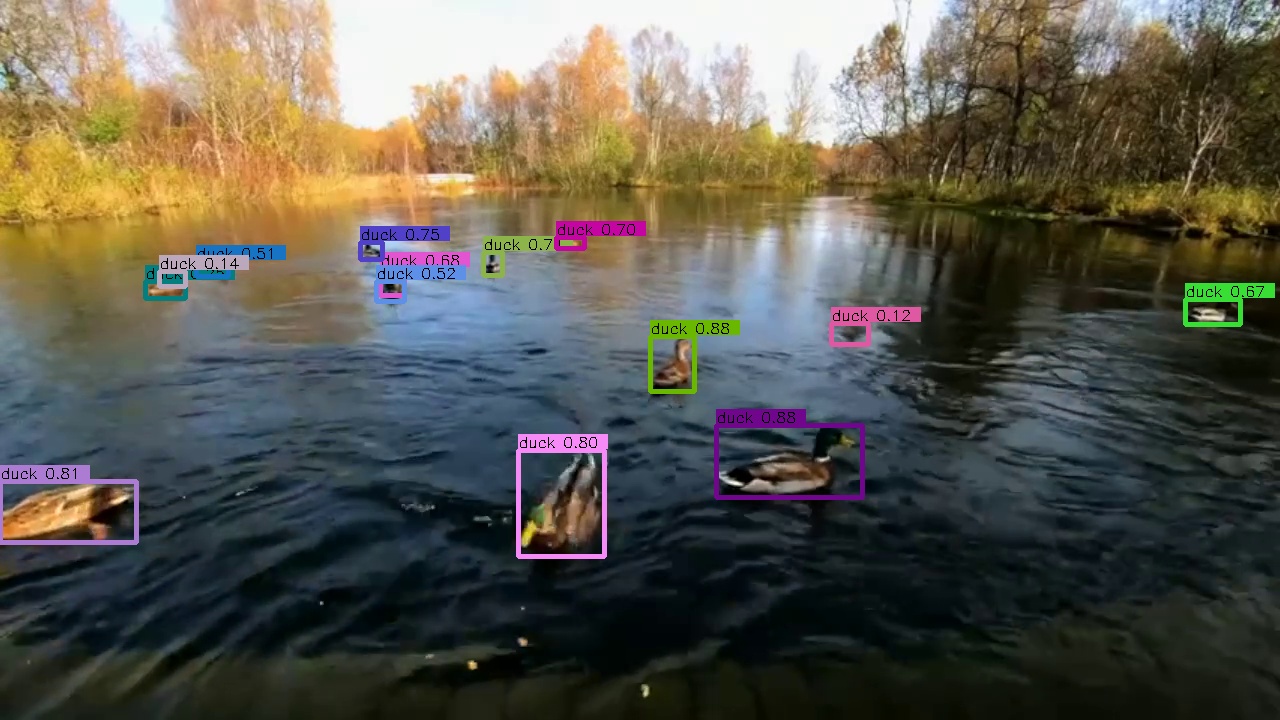} &
        \includegraphics[width=0.2\linewidth]{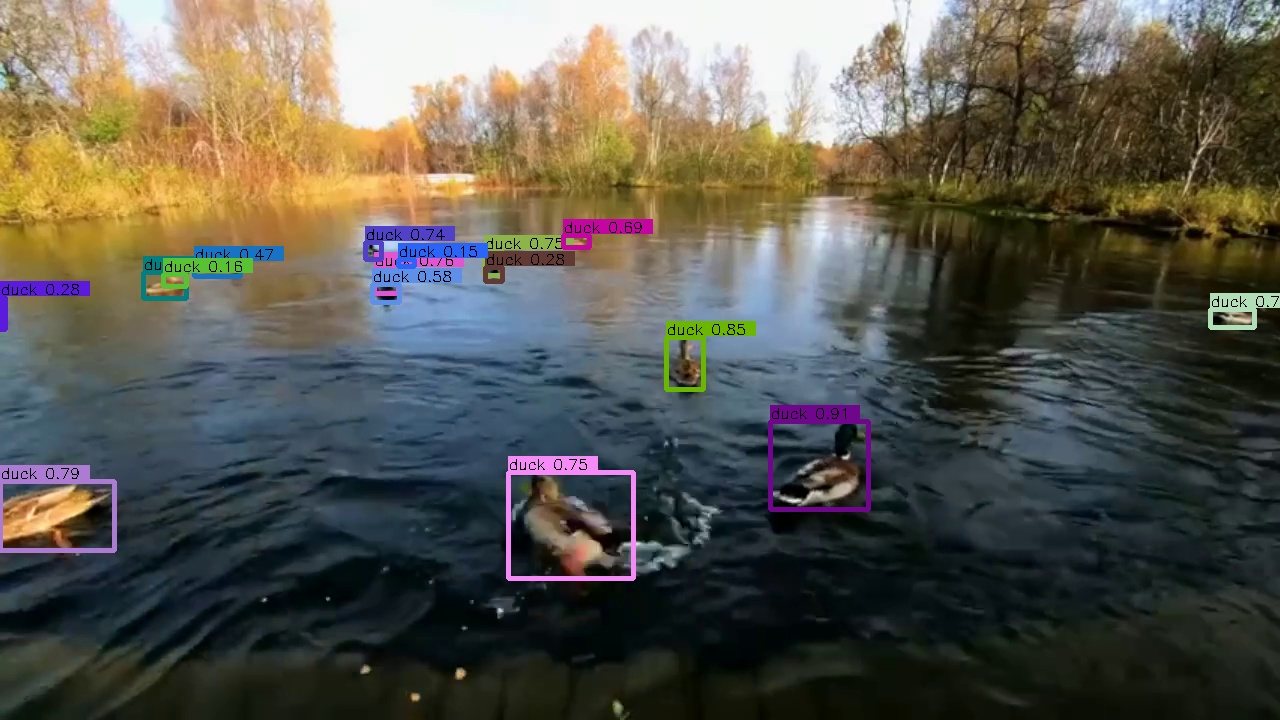} &
        \includegraphics[width=0.2\linewidth]{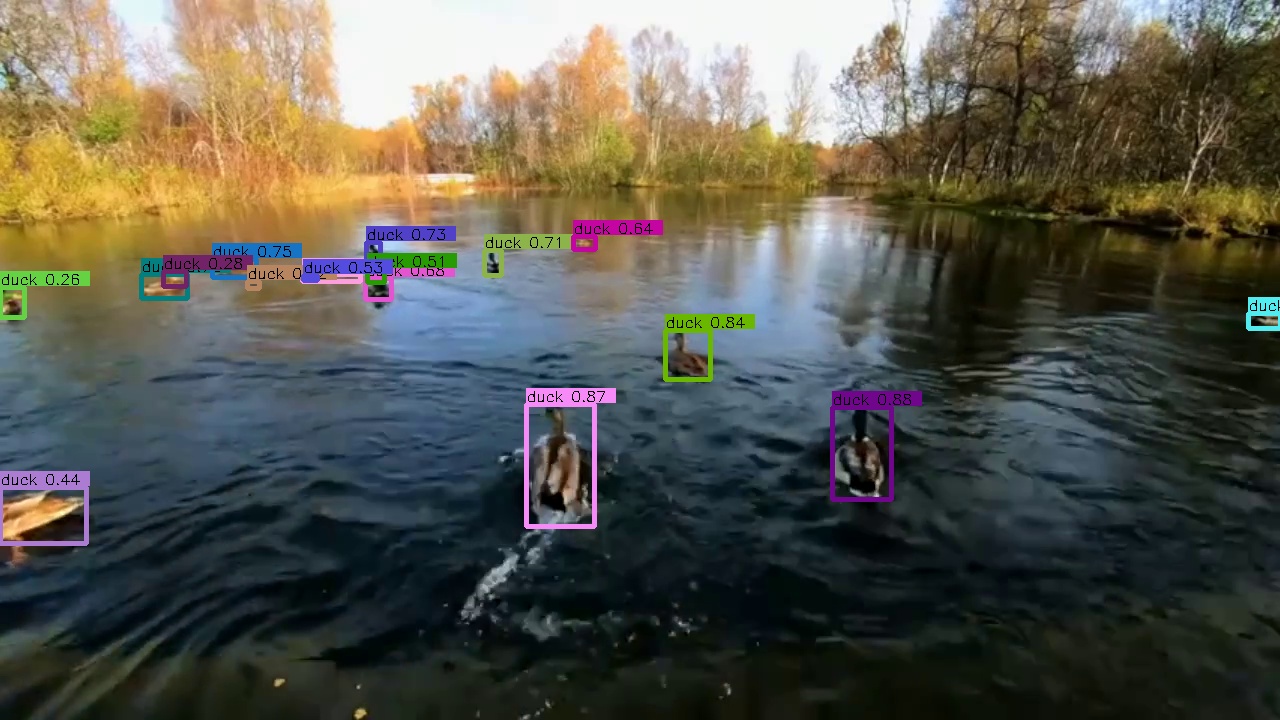} \\

        \bottomrule
    \end{tabular}}
          \caption{\textbf{Qualitative results of Open-Vocabulary Tracking}. We condition SLAck on text prompts unseen during testing and successfully track the corresponding objects in the videos. The bounding box colour depicts the object's identity. We choose random internet videos to test our algorithm on diverse real-world scenarios. Best viewed digitally.}
    \label{fig:qualitative}
\end{figure*}

\subsection{Inference Detail}
During inference, the shorter side of images is resized to 800 pixels. We set all detection-related hyper-parameters such as test score threshold and NMS settings, the same as during the training. We set the Sinkhorn iteration of exit to 100 and use a matching threshold of 0.2 ($match\_score\_thr$). A memory queue lasting 10 seconds ($memo\_length$) retains semantic, location, and appearance encodings of all objects in the tracklets, with objects expiring after a 10-second inactivity period. We provide a detailed tracking algorithm pseudo-code in Algo.~\ref{alg:track}.

Compared to previous tracking algorithms~\cite{QDTrack,TETer, OVTrack,bytetrack,OC-SORT,StrongSORT}, we significantly reduce the number of hyper-parameters during inference. We only need to set two hyper-parameters which are the $match\_score\_thr$ and $memo\_length$ to decide when to match and how long the objects are stored in the tracklets. 

\subsection{Qualitative results}
The qualitative results of SLAck are shown in Fig. \ref{fig:qualitative}. We choose all the novel classes to test our tracker's ability in real-world testing scenarios.

\subsection{Limitations}
SLAck is sensitive to video frame rate changes. SLAck joint considers location, object shape changes both spatially and temporally, and the frame rate changes between training and inference can introduce a huge domain gap. For example, if the model is trained on 1 FPS video and test on 30 FPS videos, the motion or shape changes can be very different. 

Also, SLAck requires training with video labels. Despite the introduction of detection-aware training, which allows for end-to-end association training with sparse ground truth labels, the need for video annotations persists. This is particularly challenging in open-vocabulary tracking scenarios, where obtaining comprehensive video annotations is both difficult and costly. The scarcity of such annotations restricts the model's scalability and its applicability to a wider range of tracking tasks.

\begin{algorithm}
\caption{Simplified Tracking Algorithm}
\label{alg:track}
\begin{algorithmic}[1]

\State \textbf{Initialization}:
\State Initialize tracker parameters: $match\_score\_thr$, $memo\_length$, etc.
\State Initialize tracker state: $tracklets$, $fid\_tracklets$

\Procedure{UpdateMemo}{$ids$, $bboxes$, $labels$, $app_embeds$, $cls\_embeds$, $frame\_id$}
    \For{each $id$, $bbox$, $app_embed$, $cls\_embed$, $label$ in new or updated tracklets}
        \State Update or add to $tracklets$
    \EndFor
    \State Remove outdated tracklets based on $frame\_id$ and $memo\_length$
\EndProcedure

\Procedure{Match}{$bboxes$, $labels$, $app_embeds$, $cls\_embeds$, $frame\_id$, $img\_metas$}
    \State Match detections to existing objects in tracklets using the joint similarity matrix with $match\_score\_thr$
    \State Assign $ids$ to matched and newly detected tracklets
    \State \textbf{return} updated $bboxes$, $labels$, and $ids$
\EndProcedure

\State Main tracking loop:
\For{each frame}
    \State Detect objects and extract features
    \State Match detections to tracklets
    \State Update memories with current frame data
\EndFor

\end{algorithmic}
\end{algorithm}

\clearpage

\end{document}